\documentclass[11pt]{article}

\usepackage[final]{acl}

\usepackage{microtype}
\usepackage{graphicx}
\usepackage{subfigure}
\usepackage{booktabs} 
\usepackage{times}
\usepackage{xcolor}
\usepackage{siunitx}
\usepackage{tcolorbox}
\usepackage{tikz}
\usepackage{listings}
\usepackage{color}
\usepackage{latexsym}
\usepackage{multirow}
\usepackage{colortbl}
\usepackage{amsmath}
\usepackage{amssymb}
\usepackage{bbding}
\usepackage{booktabs}
\usepackage{dashrule}
\usepackage{arydshln}
\usetikzlibrary{positioning}
\usepackage{times}
\usepackage{latexsym}

\usepackage[T1]{fontenc}

\usepackage[utf8]{inputenc}

\usepackage{microtype}

\usepackage{inconsolata}

\usepackage{graphicx}

\title{Quantifying and Improving the Robustness of Retrieval-Augmented Language Models Against Spurious Features in Grounding Data}

\author{
 \textbf{Shiping Yang\textsuperscript{1 2}\thanks{Work done during an internship at Microsoft}},
 \textbf{Jie Wu\textsuperscript{3}\thanks{Correspondence to: \href{mailto:email@domain}{jiewu\_ecnu@hotmail.com}}},
 \textbf{Wenbiao Ding\textsuperscript{2}},
 \textbf{Ning Wu\textsuperscript{2}},
 \textbf{Shining Liang\textsuperscript{2}},\\
 \textbf{Ming Gong\textsuperscript{3}}, 
 \textbf{Hongzhi Li\textsuperscript{4}},
 \textbf{Hengyuan Zhang\textsuperscript{5}}, 
 \textbf{Angel X. Chang\textsuperscript{1 6}},
 \textbf{Dongmei Zhang\textsuperscript{2}}
\\
 \textsuperscript{1}Simon Fraser University  \ \
 \textsuperscript{2}Microsoft  \ \
 \textsuperscript{3}Atlassian  \ \
 \textsuperscript{4}Tongji University \\
 \textsuperscript{5}The University of Hong Kong \ \
 \textsuperscript{6}Canada-CIFAR AI Chair, Amii \ \
 \\
}

\begin{document}
\maketitle
\begin{abstract}
Robustness has become a critical attribute for the deployment of RAG systems in real-world applications. Existing research focuses on robustness to explicit noise (e.g., document semantics) but overlooks implicit noise (spurious features). Moreover, previous studies on spurious features in LLMs are limited to specific types (e.g., formats) and narrow scenarios (e.g., ICL). In this work, we identify and study spurious features in the RAG paradigm, a robustness issue caused by the sensitivity of LLMs to semantic-agnostic features. We then propose a novel framework, \textit{SURE}, to empirically quantify the robustness of RALMs against spurious features. Beyond providing a comprehensive taxonomy and metrics for evaluation, the framework's data synthesis pipeline facilitates training-based strategies to improve robustness.
Further analysis suggests that spurious features are a widespread and challenging problem in the field of RAG. Our code is available at \url{https://github.com/maybenotime/RAG-SpuriousFeatures}.
\end{abstract}

\section{Introduction}
\label{sec:intro}


Retrieval-Augmented Generation (RAG) has emerged as a promising paradigm to mitigate LLMs hallucinations~\cite{ragsurvey,yang2023hallu}, integrating relevant external knowledge to improve the factuality and trustworthiness of LLM-generated outputs~\cite{ragtrustworthiness}.
However, Retrieval-Augmented Language Models (RALMs) still face substantial robustness issue due to the presence of noise in retrieved documents~\cite{recall,li2024dalk}. 

\begin{figure}[h]
    \centering
    \resizebox{0.48\textwidth}{!}{\includegraphics{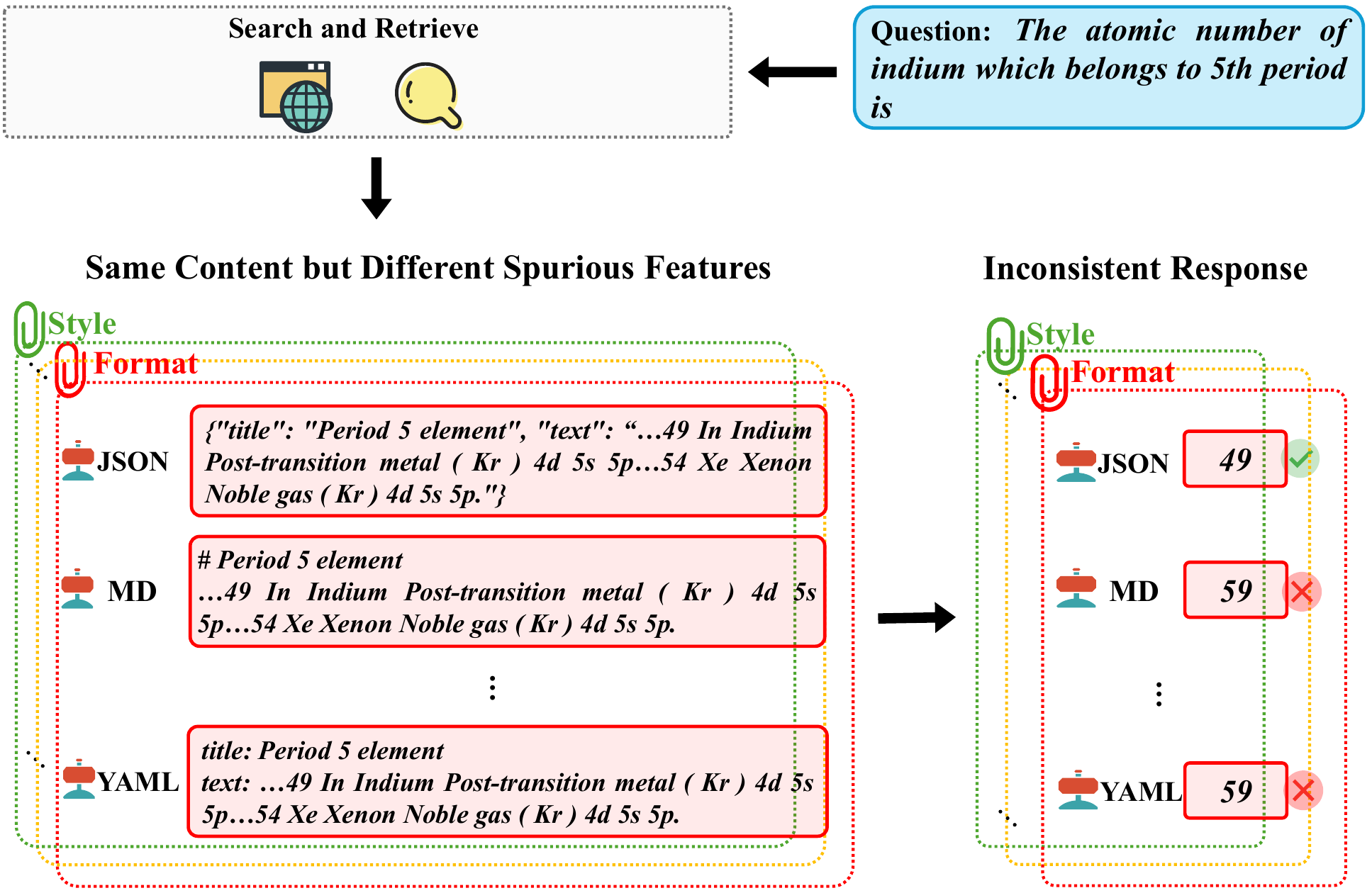}}
    \caption{An example from the \textit{SURE\_Wiki} dataset (Sec. \ref{sec:experiments}), illustrating the sensitivity of RAG systems to spurious features within grounding data. The original retrieved document is fed into the LLMs in different formats, leading to inconsistent responses.}
    \label{fig:example}
\end{figure}

Recent research aims to explore the characteristics of grounding data that influence the robustness of RAG systems~\cite{powerofnoise}. These studies examine various factors, including the type~\cite{pandora}, number~\cite{retrievalmeetslong}, and position of documents~\cite{lostinmiddle} within the prompt context.
However, previous analyses primarily focus on explicit noise that significantly alter the semantic information (causal features) of grounding data~\cite{clasheval,powerofnoise}, while neglecting implicit noise (spurious features) that introduce semantic-agnostic modifications. This limitation extends to existing evaluation benchmarks, which simulate complex noise scenarios to assess the robustness of RAG systems~\cite{benchmarkingrobustness,pandora}, yet lack available benchmarks and metrics to measure the robustness of RALMs\footnote{We focuses on the generation end of RAG systems (i.e., RALMs), and hereafter RAG systems refer to RALMs.} against spurious features. 


Contemporary RAG systems typically employ production-level retrievers, such as Bing and Google, to retrieve relevant information from the internet. Unlike a single corpus, the internet encompasses diverse data with distinct features. For any given query, there may exist numerous golden documents that contain the correct answer but differ in style, format, or other attributes. As shown in Figure \ref{fig:example}, we have observed that LLMs may fail to consistently derive the correct answer from golden documents with different formats.
A similar phenomenon is reported in ~\citet{quantifyingsensitivity} and ~\citet{formatimpact}, which demonstrate that LLMs are extremely sensitive to the format of prompts (i.e., spurious features).
However, spurious features in the RAG paradigm have not been empirically and systematically studied, nor have effective mitigation strategies been proposed.

In this work, we first propose a novel framework, \textit{SURE}, for automating the process of robustness evaluation. This framework follows a \textit{perturb-then-evaluate} approach, offering great scalability. In \textit{SURE}, automated perturbations are applied to the original instances to inject the corresponding spurious features. The perturbed instances are then examined to ensure that the causal features remain intact. After these steps, we employ tailored metrics to quantify the robustness of RALMs against spurious features. 
To enable more efficient evaluation, we select the most challenging instances to create a lighter benchmark, \textit{SIG\_Wiki}. Extensive evaluations across diverse LLMs and methods indicate that maintaining robustness against spurious features is a widespread and challenging issue. To address this, we introduce two training-based strategies to mitigate the lack of robustness caused by spurious features, leveraging the synthetic data generated by our \textit{SURE} framework.

Our contribution can be summarized as follows:
\textbf{1)} To the best of our knowledge, this is the first comprehensive study to evaluate spurious features from RAG perspective. We propose a novel evaluation framework, \textit{SURE}, to assess the robustness of RALMs against spurious features, which includes a comprehensive taxonomy, tailored metrics, and a data synthesis pipeline. 
\textbf{2)} Using the synthetic dataset generated by the \textit{SURE} framework, we curate a lightweight yet challenging evaluation benchmark, \textit{SIG\_Wiki}, and introduce two effective training-based strategies to improve the robustness of RALMs.
\textbf{3)} Further analysis offer valuable insights for future research. For example, we found that not every spurious features is harmful and they can even be beneficial sometimes.

\section{Related Work}
\label{sec:related}
\subsection{Robustness Evaluation of RAG Systems}
\label{sec:related_a}
The retrieved contexts inevitably contains noise in addition to desirable knowledge, which may mislead LLMs to produce an incorrect response~\cite{externalinfluence,double-edgedsword}. Previous works have explored automated evaluation frameworks to assess the robustness of RAG systems in various settings~\cite{benchmarkingrobustness,powerofnoise}. 
\citet{pandora} provided a detailed taxonomy of noise documents to further simulate the complexity of real-world scenarios and highlighted the potential positive effects of certain types of noise.

While these studies have identified several explicit noises that affect the robustness of RAG systems, they predominantly overlook implicit noises. Even some works evaluate the influence of implicit noise, they are often limited to specific types, such as typos~\cite{cho-etal-2024-typos} and formatting~\cite{htmlrag}.
In this work, we comprehensively study semantic-agnostic noises (i.e, spurious features) in RAG systems.


\subsection{Prompt Sensitivity of LLMs}
\label{sec:related_b}
Prompts are instructions provided to an LLM to perform specific tasks automatically and ensure desired qualities in the generated output. However, it is known that current LLMs are sensitive to the features of input prompts~\cite{promptrobustnessbench}. This sensitivity poses challenges for researchers attempting to evaluate the model's performance accurately and precisely~\cite{prosa}.  
Beyond causal features that significantly influence the meaning of prompts~\cite{wei2022chain,roleplayprompting}, existing works have demonstrated that LLMs are highly sensitive to spurious features~\cite{quantifyingsensitivity} in non-RAG scenarios, e.g, prompt formatting~\cite{formatimpact}, language style~\cite{emotionprompt}, the order of options~\cite{ordersensitivity}. 

In contrast, our work focuses on spurious features within the context of RAG.
Unlike non-RAG scenarios (e.g., ICL), where spurious features appear in user-provided instructions and remain static across queries, in RAG they occur in retrieved documents and vary dynamically for each query.




\section{Proposed Framework}
\label{sec:framework}
 \begin{figure*}[h]
    \centering
    \resizebox{0.99\textwidth}{!}{\includegraphics{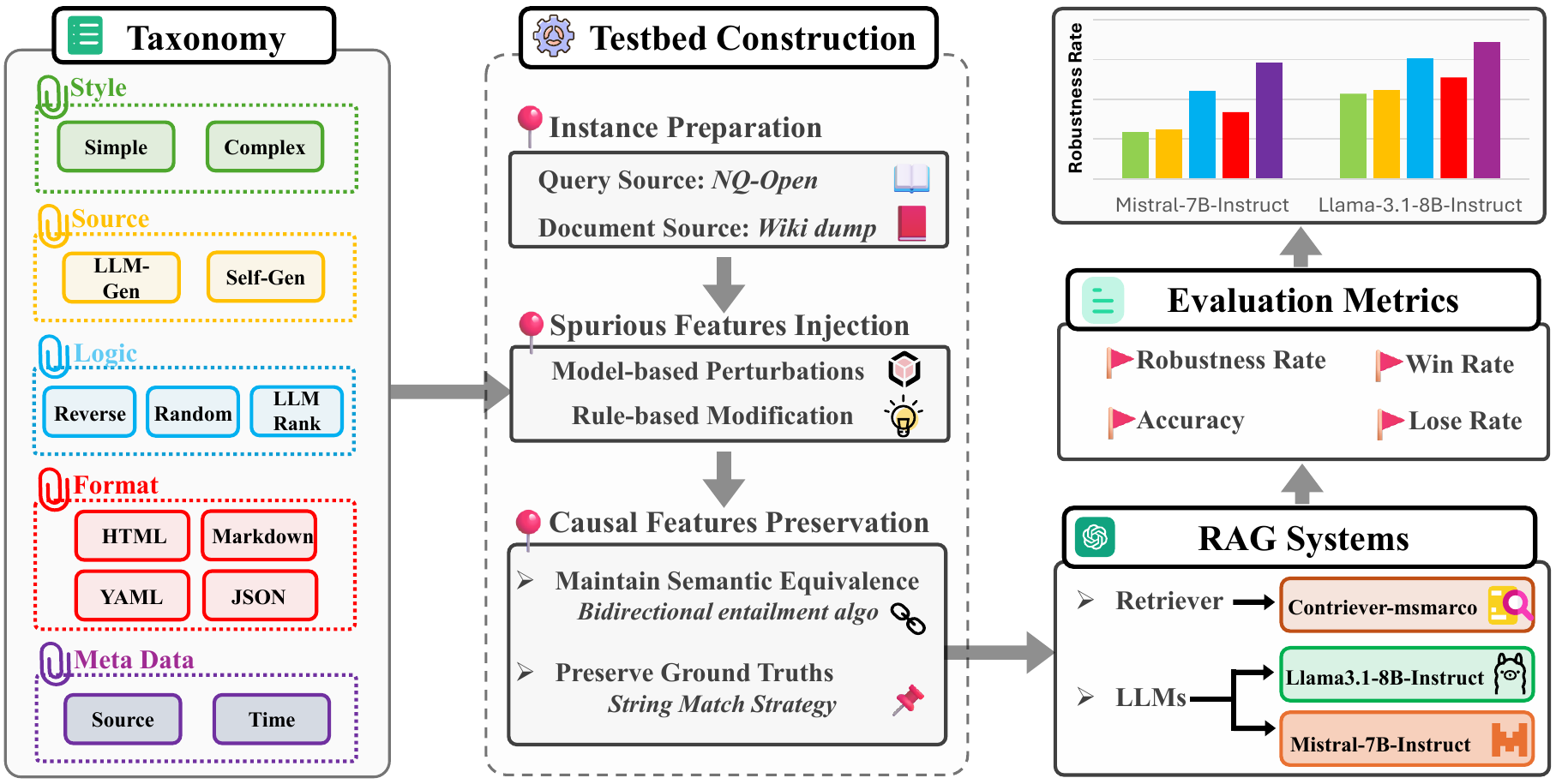}}
    \caption{Overview of our SURE framework. We provide a \textit{Comprehensive Taxonomy} that includes five types of spurious features, further divided into 13 subtypes of perturbations (left section). To construct the testbed, we prepare raw instances initially and then synthesize the modified instances through a workflow consisting of \textit{Spurious Features Injection} and \textit{Causal Features Preservation} (middle section). By applying carefully tailored metrics for \textit{Robustness Evaluation}, we quantify the robustness of target RAG systems (right section).}
    \label{fig:SURE_overview}
\end{figure*}

In this section, we detail our proposed evaluation framework, \textit{SURE} (\textbf{S}purious Feat\textbf{U}res \textbf{R}obustness \textbf{E}valuation), which designed specifically for assessing the robustness of RALMs against spurious features in grounding data. As illustrated in Figure \ref{fig:SURE_overview}, this framework comprise four components: \textbf{1)} \textit{Comprehensive Taxonomy.} We identify and define five common types of spurious features in RAG scenarios.
\textbf{2)} \textit{Spurious Features Injection.} We design a data synthesis pipeline to automate the injection of spurious features, utilizing both model-based and rule-based methods to construct counterparts of the original document with varying spurious features.
\textbf{3)} \textit{Causal Features Preservation.} We employ a bidirectional entailment algorithm and a string matching strategy to ensure that the causal features of grounding data remain unchanged.
\textbf{4)} \textit{Robustness Evaluation.} We introduce three metrics (Win Rate, Lose Rate, and Robustness Rate) to facilitate fine-grained, instance-level evaluation.

\subsection{Problem Formulation}
\label{section:problem formulation}
Given a query \(q\) , the retriever \(R\)  returns a list of relevant documents from a corpus $D = \{ d_i \}_{i=1}^{N}$.
The relevance between document \(d\) and query \(q\) can be measured by various methods. In this work, we use a BERT-based dense retriever to obtain the embedding of query and documents, respectively. The relevance score is calculated by computing their dot-product similarity.
Then, the Top-k documents with the highest similarity scores are retrieved:
\begin{equation}
D_{\text{retrieve}} = \text{argtop-}k \left\{ s(q, d_i) \mid d_i \in D \right\}.
\end{equation}
To formally quantify the robustness of RAG systems against spurious features, we define the input prompt for the LLM-based reader as $P= (I,G,Q)$, where \(I\) represents instruction, \(G\) refers to the grounding data, constituted by a subset of \(D_{\text{retrieve}}\), and \(Q\)  is the query. A perturbation is introduced to investigate the impact of spurious features by applying a semantic-agnostic modification to the original grounding data, while preserving its causal features. We define \( g(.) \) to automate this process, transforming \( G \) to \( g(G) \) and producing a counterpart \( \hat{P} = (I,g(G),Q) \). The outputs of LLM-based reader for \(P\) and \(\hat{P}\) are compared to evaluate the impact of the introduced perturbation:
\begin{equation}
y = \text{LLM}(P), \quad \hat{y} = \text{LLM}(\hat{P}).
\end{equation}

\subsection{Taxonomy of Spurious Features}
We develop a comprehensive taxonomy of spurious features, informed by our preliminary experiments (see Appendix~\ref{appendix: preliminary}) and insights from prior research.
The five types of spurious features and their corresponding perturbations are detailed below.

\paragraph{Style Perturbations}
The same content can be expressed in different styles, using varying tones, words and sentence structures. As shown in Appendix~\ref{subsec:pre_exp}, LLMs exhibit biases towards readability-related features. 
Similarly, for humans, the readability of a text can significantly influence its accessibility to the audience~\cite{css}.
Therefore, we define two perturbations from the perspective of readability style: \textbf{Simple} and \textbf{Complex}. The former simplifies the grounding data by using basic vocabulary and simple sentence structure, while the latter employs professional vocabulary and a formal academic tone to complex the documents.

\paragraph{Source Perturbations}
Recent studies have shown that neural retrievers are biased towards LLM-generated content, leading to the marginalization of human-authored content~\cite{biasedretriever,spiralofsilence}. Moreover, our preliminary experiments demonstrate that LLMs are biased towards the Perplexity (PPL) of text. Thus, we define two types of source perturbations: \textbf{LLM-generated} and \textbf{Self-generated}. Specifically, the LLM-generated perturbation paraphrases the original document using a powerful LLM, while the self-generated perturbation employs the same backbone model used as the generator in the RAG system.

\paragraph{Logic Perturbations}
In RAG systems, documents are often segmented into multiple chunks and may be retrieved in varying orders. Here, we simulate scenarios where the intrinsic logical flaw is disrupted by three different perturbations: \textbf{Random}, \textbf{Reverse}, and \textbf{LLM-reranked}, each representing a distinct sentence ordering strategy.

\paragraph{Format Perturbations}
The internet contains various data formats, including \textbf{HTML}, \textbf{Markdown}, \textbf{YAML} and \textbf{JSON}. These formats are usually processed into plain text before being fed to LLMs. To mitigate the loss of structural information during this process, some RAG studies propose using the original format, rather than plain text, to augment the generation~\cite{htmlrag}. However, as highlighted in previous research, the prompt format is recognized as a spurious feature that can significantly impact model performance~\cite{quantifyingsensitivity,formatimpact}. Therefore, we perturb the original document with four common formats to explore the impact of grounding data format.

\paragraph{Metadata Perturbations}
Metadata is often included in the HTML results returned by search engines. In our framework, we focus on two types: \textbf{Timestamp} and \textbf{Data source}. The timestamp marks when the data was created, and the data source indicates its origin \footnote{Timestamp may serve as causal features in time-sensitive tasks. In our experiments, the NQ dataset generally does not contain time-sensitive queries.}.
For timestamp perturbations, \textit{pre} and \textit{post} denote whether the timestamp is before or after the LLM’s knowledge cutoff date. For data source perturbations, \textit{wiki} and \textit{twitter} represent the domains of the URLs.

\subsection{Spurious Features Injection}
The automation of spurious features injection is essential for automating the entire evaluation framework. We detail the process of collecting the original instances and describe how the automated perturbation was implemented.

\paragraph{Instance Preparation}
An instance is the dynamic component of the prompt \( P \), consisting of a query \( Q \) and grounding data \( G \). 
To construct the original instances, we first select 1,000 queries from the NQ-open dataset based on the close-book QA results of \textit{Mistral-7B-Instruct-v0.3}.
This subset includes 500 queries that can be answered directly using parametric knowledge (\textit{Known}) and 500 queries that require external knowledge for answering (\textit{Unknown}). 
For each query, we then retrieve 100 documents from the Wikipedia dump to serve as grounding data, yielding 100,000 instances for the following perturbation step. 


\paragraph{Automated Perturbation}
As introduced in Section \ref{section:problem formulation}, the perturbation \( g(.) \) injects spurious features by modifying the grounding data. For style and source perturbations, \( g(.) \) is implemented using an LLM\footnote{Unless otherwise specified, all model-based \( g(.) \) are implemented using \textit{Llama-3.1-70B-Instruct}.} prompted by carefully crafted guidelines to modify the raw document, producing counterparts of the original instances. For logic and format perturbations, we develop \( g(.) \) as a heuristic method based on a set of predefined rules. To simulate real-world metadata, we first synthesize pseudo Wikipedia or Twitter links for the raw instances, and then organize them into HTML format using a rule-based \( g(.) \). Further details for automated perturbation are provided in Appendix \ref{appendix: spurious_features_injection}.



\subsection{Causal Features Preservation}
To eliminate the effect of causal features, it is essential to follow the principle of controlled experiments by keeping causal features constant while systematically manipulating spurious features. This approach isolates the impact of spurious features from that of causal features, enabling an accurate quantification of robustness against spurious features. In our framework, we introduce two methods to ensure the stability of causal features in the grounding data. Implementation details can be found in Appendix \ref{appendix: causal_features_preservation}.

\paragraph{Maintain Semantic Equivalence}
For models capable of following human instructions, we directly instruct them to maintain semantic equivalence when injecting spurious features. Nonetheless, it's impossible to completely avoid semantic shift during the perturbation process. To ensure the semantic consistency before and after introducing perturbation, we employ a bidirectional entailment algorithm to filter out instance pairs (raw instance, perturbed instance) with semantic inequivalence. Specifically, for document \( G \) and its modified counterpart \( g(G) \), we use a Natural Language Inference (NLI) system to detect whether the latter can be inferred from the former, and vice versa. The NLI system classifies predictions into one of: \textit{entailment}, \textit{neutral}, \textit{contradiction}. We compute both directions, and the algorithm returns \textit{equivalent} if and only if both directions are predicted as entailment.

In general, this algorithm can be implemented by any NLI system. However, in our case, the concatenation of \( G \) and \( g(G) \) sometimes exceeds the context limitation of a Bert-based NLI model. Hence, we apply an LLM-based NLI system \footnote{\citet{semanticNature} confirms the effectiveness of the LLM-based NLI system through human annotation, demonstrating that its performance is on par with the DeBERTa-large model used in \citet{semanticICLR}. } to implement the bidirectional entailment algorithm. 

\paragraph{Preserve Ground Truths}
While semantic equivalence protects causal features to the greatest extent, the perturbation may lead to the correct answer being paraphrased into an alias (e.g., "President Roosevelt" to "Roosevelt"). 
To address this issue, we employ a simple string-matching strategy to filter out documents that have undergone unexpected modifications.

\subsection{Robustness Evaluation}
\begin{figure}[h]
    \centering
    \resizebox{0.48\textwidth}{!}{\includegraphics{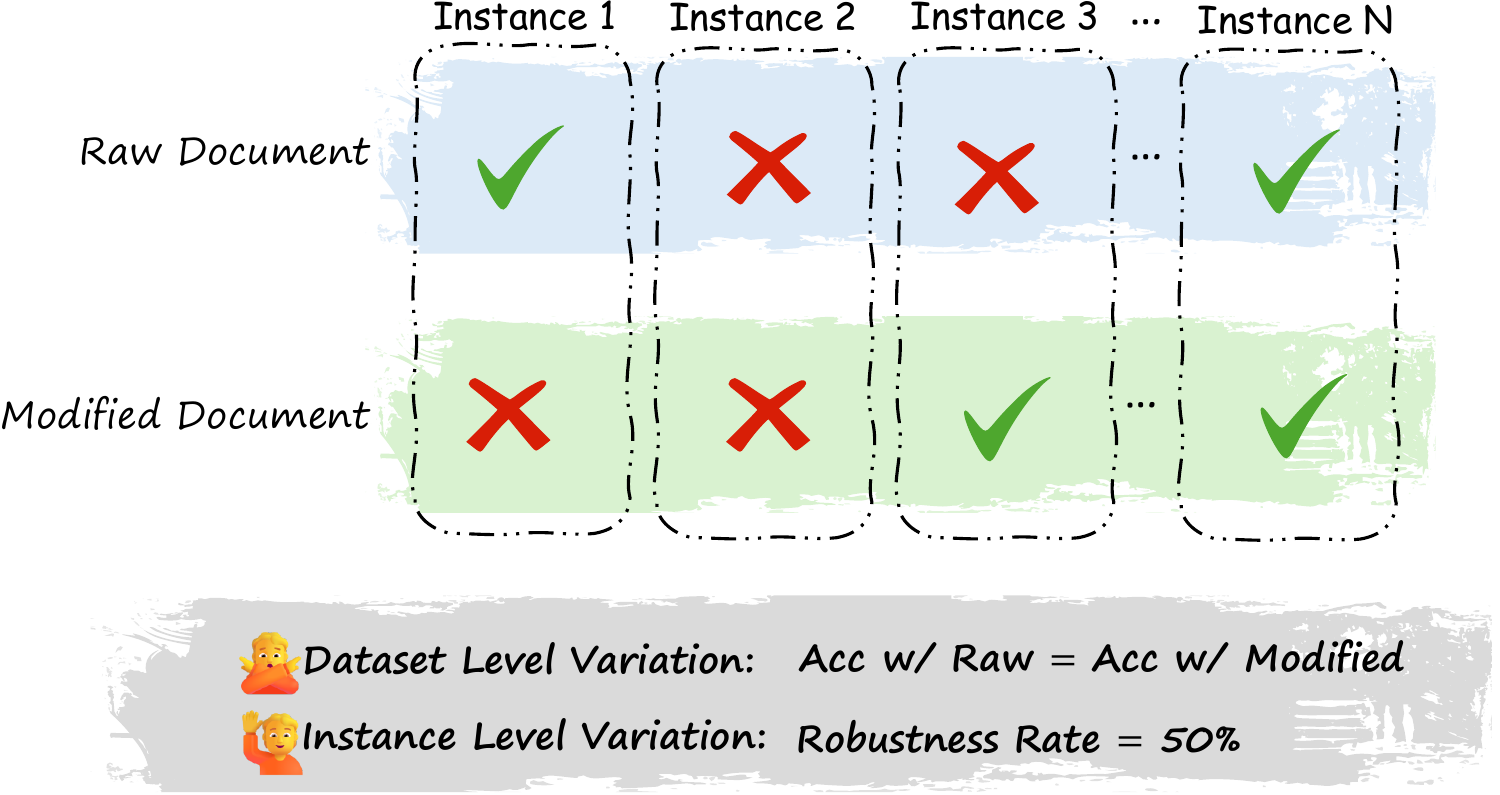}}
    \caption{A comparison of dataset-level metric (Acc) and instance-level metric (RR) for robustness evaluation. \CheckmarkBold and \XSolidBrush indicate the correctness of responses. In this example, RR captures instance-level unrobustness, while Acc overlooks RALMs' sensitivity to spurious features within documents.  }
    \label{fig:robustness_rate}
\end{figure}

 We employ an evaluation method \(Y(.)\), in line with ~\citet{lostinmiddle,powerofnoise}, to measure the correctness of responses generated by RAG systems. This approach checks whether any of the correct answers is contained within the response produced by the LLM and then derives a binary label. Previous researches use accuracy as the primary metric and report it at dataset level to assess the robustness of RALMs, which is quantified by calculating the variations in the models' accuracy across different types of noise. However, dataset-level metrics has certain limitations, as it may fail to capture fine-grained variations that occur at the instance level.
As shown in Figure \ref{fig:robustness_rate}, RALMs may appear robust at dataset-level evaluations but exhibit significant sensitivity at the instance level.

To quantify whether a RAG system is robust and unbiased at the instance level, we assign a ternary label to each instance by comparing the correctness of the LLM's response before and after introducing the perturbation. This comparison process can be formulated as $C = Y(y_i) - Y(\hat{y_i})$, where \(C\) lies in the set $(-1,0,1)$. Based on the comparison outcomes, we define three metrics: \textbf{Robustness Rate (RR)}, \textbf{Win Rate (WR)}, and \textbf{Lose Rate (LR)}. The RR is calculated as follows:

\begin{equation}
\text{RR} = \frac{1}{N} \sum_{i=1}^{N} \mathbb{I}(C == 0)
\end{equation}

where \(N\) is the total number of instances in the dataset; \(y_i\) and \(\hat{y_i}\) represent the outputs of LLM for the original and perturbed instances. RR measures the proportion of instances where the RALM's answer remains consistent (0) before and after introducing the perturbation. Similarly, WR and LR quantify the proportions of instances where the correctness of the RALM's response changes after the perturbation, either from incorrect to correct ($C==-1$) or from correct to incorrect ($C==1$).

\section{Experiments}
\label{sec:experiments}
\begin{table*}[h]
\centering
\vspace{-0.2cm}
\renewcommand{\arraystretch}{1.1}
\resizebox{1.0\textwidth}{!}{
\begin{tabular}{clcccccccccccccccc}
\toprule[2pt]
                     \multicolumn{18}{c}{\textit{Mistral-7B-Instruct-v0.3}}                      \\ \addlinespace[2pt]
\cline{1-18} \addlinespace[2pt]
\multirow{2}{*}{\textbf{Taxonomy}}       & \multirow{2}{*}{\textbf{Perturbations}}             & \multicolumn{5}{c}{\textbf{Known-Golden}} & \multicolumn{5}{c}{\textbf{Known-Noise}} & \multicolumn{5}{c}{\textbf{Unknown-Golden}} & \multicolumn{1}{c}{\textbf{U-N}}  \\
\cmidrule(lr){3-7} \cmidrule(lr){8-12} \cmidrule(lr){13-17} \cmidrule(lr){18-18}
&                                & LR & RR & WR & Org & Acc & LR & RR & WR & Org & Acc & LR & RR & WR & Org & Acc & RR \\ \addlinespace[2pt]
\hline \addlinespace[2pt]
\multirow{2}{*}{Style}     & Simple & 7.33        & 85.00        & \textbf{7.67}   & \multirow{2}{*}{73.02}      & 73.37     & 4.45      &  91.64     & 3.90  &  \multirow{2}{*}{10.82}    & 10.28         & 7.87        & 82.95        & \textbf{9.18}        & \multirow{2}{*}{56.31}   &  57.62   &  98.76      \\
                            & Complex & 6.05       &  87.42      & \textbf{6.53}   &    
 & 73.50     & 3.85      & 92.03      & \textbf{4.12}  &     & 11.10     & 6.90        & 85.92        & \textbf{7.17}  &    & 56.58         & 98.82       \\ \addlinespace[2pt]  \hdashline[1pt/1pt]  \addlinespace[2pt]
  
\multirow{2}{*}{Source}       & LLM-Generated & 5.91       & 87.62       &  \textbf{6.47}  &  \multirow{2}{*}{71.81} & 72.36    & 3.57      & 92.27      & \textbf{4.16}   &  \multirow{2}{*}{10.79}   & 11.38   & 6.41         & 86.52         & \textbf{7.06}   & \multirow{2}{*}{54.46}
  & 55.11       &  98.75         \\
                            & Self-Generated & 6.30       &  87.06      &  \textbf{6.64}   &     & 72.15     & 3.94      & 92.02      & \textbf{4.04}   &     &  10.89  & 6.26        & 86.80        & \textbf{6.94}     &   &  55.14     & 98.77          \\
\addlinespace[2pt]  \hdashline[1pt/1pt]  \addlinespace[2pt]  
\multirow{3}{*}{Logic}       & Reverse &  5.44      & 89.34       & 5.22  & \multirow{3}{*}{69.91}  & 69.69      & 2.99      &  94.10     & 2.92   &  \multirow{3}{*}{11.77}  & 11.70  & 5.97        & 88.54       & 5.49    & \multirow{3}{*}{50.26}   & 49.79       & 99.04        \\
                            & Random &  4.47      &  91.87      & 3.66   &  & 69.10    & 2.43      & 95.15      & 2.42   &    & 11.76   & 4.18        & 91.44        & \textbf{4.38}        &      &  50.46    & 99.27        \\
                            & LLM-Ranked & 3.52       & 93.15       & 3.33  &    & 69.72     & 2.07      & 95.84      & \textbf{2.09}   &    & 11.79   & 3.57        & 92.89        & 3.54    &    &  50.24      &  99.30     \\
\addlinespace[2pt]  \hdashline[1pt/1pt]  \addlinespace[2pt]     
\multirow{4}{*}{Format}       & JSON & 7.96       & 88.53        & 3.51    & \multirow{4}{*}{70.81}   & 66.35   & 5.15      & 92.68      & 2.17  & \multirow{4}{*}{10.98}  & 8.00  & 6.95        & 88.92        & 4.13   &  \multirow{4}{*}{53.32}   &  50.50      & 99.02          \\
                            & HTML & 9.30       & 87.03       & 3.67    &     & 65.18   & 5.89      & 92.36      & 1.74  &    & 6.83   & 8.36        & 87.39        & 4.25      &   &   49.22     & 99.01     \\
                            & YAML & 4.75       &  90.90      &  4.35   &    & 70.41  & 3.88      & 93.24      & 2.87   &    & 9.97   & 5.05        & 90.53        & 4.42      &    & 52.69       & 99.06        \\
                            & Markdown & 3.98       & 92.49      & 3.53  &   &  70.36    & 2.91      & 94.36      & 2.72    &    & 10.79  & 4.11        & 92.59        & 3.31    &    & 52.52      & 99.15          \\
\addlinespace[2pt]  \hdashline[1pt/1pt]  \addlinespace[2pt]    
\multirow{4}{*}{Metadata}      & Timestamp (pre) &  2.62      & 94.90       & 2.48  & \multirow{4}{*}{65.04}   & 64.90    & 1.28      & 97.61      & 1.11    &  \multirow{4}{*}{6.83}  & 6.66     &  3.15       & 94.45        & 2.40     & \multirow{4}{*}{48.08}     & 47.33       & 99.67         \\
                            & Timestamp (post) & 2.74       & 94.87       & 2.40   &   & 64.70     & 1.16      & 97.63      & \textbf{1.21}  &    &  6.88   & 3.45        & 94.41        & 2.14     &   &  46.77       & 99.68             \\    
                             & Datasource (wiki) & 3.78      & 92.31       & \textbf{3.91}   &    & 65.17      & 1.5      & 96.66      & \textbf{1.84}    &     &  7.16   & 3.69        & 92.95        & 3.36        &       &  47.76       & 99.48          \\
                            & Datasource (twitter) & 2.68       & 93.59       & \textbf{3.73}   &     &  66.08    & 1.3      & 97.22      & \textbf{1.48}   &    & 7.00    & 2.04        & 94.90        & \textbf{3.06}    &       &  49.10      & 99.59          \\
\bottomrule[2pt]

\end{tabular}}
\caption{Robustness evaluation results of \textit{Mistral-7B-Instruct-v0.3} on the \textit{SURE\_Wiki} dataset. \textit{Org} indicates the accuracy on original instances, while \textit{Acc} refers to the accuracy after introducing perturbations. We use \textbf{Bold} to mark the WR values that are higher than the LR, suggesting that the perturbation is beneficial.} 
\label{tab:mistal_7b_sure_results}
\end{table*}

In this section, we assess the robustness of RAG systems to spurious features by evaluating them on their most popular application---the Question Answering (QA) task, following the standard "retrieve-read" setting of the RAG paradigm.

\subsection{Experimental Setup}
\label{sec:experimentalsetup}
\paragraph{Datasets}
Through the steps of \textbf{spurious features injection} and \textbf{causal features preservation}, we derive the final dataset available for robustness evaluation: \textit{SURE\_Wiki}. The queries are drawn from the NQ-open dataset~\cite{nq-open}, while our data source is English Wikipedia dump. 

\paragraph{Models}
We test two representative LLMs in our main experiments: \textit{Mistral-7B-Instruct-v0.3} and \textit{Llama-3.1-8B-Instruct}. Further implementation details are included in Appendix \ref{appendix: setup details}.

\begin{table}[ht]
\centering
\resizebox{0.48\textwidth}{!}{
\begin{tabular}{lrrrrr}
\toprule[2pt]
\textbf{} & \textbf{K-G} & \textbf{K-N} & \textbf{U-G} & \textbf{U-N} & \textbf{Total} \\ \addlinespace[2pt]
\hline \addlinespace[2pt]
\textbf{Style} & 7766 & 31152 & 2593 & 37692 & 79203 \\
\textbf{Source} & 9249 & 32435 & 3228 & 39101 & 84013 \\
\textbf{Logic} & 9724 & 35537 & 3587 & 41990 & 90838 \\
\textbf{Format} & 11037 & 38018 & 4141 & 45518 & 98714 \\
\textbf{Meta} & 11104 & 38018 & 4255 & 45420 & 98797 \\
\bottomrule[2pt]
\end{tabular}}
\caption{Statistics of the \textit{SURE\_Wiki} dataset for \textit{Mistral-7B-Instruct-v0.3}. K-G denotes the instances composed of (\textit{Known} query, \textit{Golden} Document), while U-N refers to the instances consisting of (\textit{Unknown} query, \textit{Noise} Document). The values represents the number of instance pairs for each type of perturbations within the category of spurious features.}
\label{tab:data_summary_mistral}
\end{table}

\subsection{Experimental Results}
To further analyze spurious features, we divide \textit{SURE\_Wiki} into four subsets based on the categories of queries and documents within each instance. A query is labeled as \textit{Known} if it can be correctly answered in a closed-book setting; otherwise, it is labeled as \textit{Unknown}. Documents are categorized as \textit{Golden} or \textit{Noise} depending on whether they contain ground truths.
Notably, the distribution of the dataset is model-specific, as the classification of \textit{Known} and \textit{Unknown} queries is determined by the intrinsic knowledge of the target LLM. Table \ref{tab:data_summary_mistral} presents dataset statistics for \textit{Mistral-7B-Instruct-v0.3}, while the distribution for \textit{Llama-3.1-8B-Instruct} is shown in Appendix \ref{appendix: statistics}.

\paragraph{For Different Queries and Grounding Data}

We report the results of \textit{Mistral-7B-Instruct} and \textit{Llama-3.1-8B-Instruct} in Table \ref{tab:mistal_7b_sure_results} and Table \ref{tab:llama3_sure_results}, respectively.
For golden documents, the robustness rates of K-G and U-G are very similar for both \textit{Mistral} and \textit{Llama}, whereas their accuracy differ significantly. This suggests that, unlike robustness to explicit noise~\cite{clasheval}, \textbf{robustness against spurious features is independent of the model's internal prior knowledge}.

When tested on noise documents, the RR remains high across all spurious features, as LLMs consistently generate incorrect responses in the absence of ground truths. 
Therefore, we primarily focus on the RR results for the golden documents in the following experiments.

\paragraph{For Different Perturbations}
We observe notable differences in robustness rates across the five types of spurious features. However, within each category, the RR values for different perturbations are relatively similar. Hence, the robustness of spurious features can be estimated by averaging the RR values of their corresponding perturbations.

When further comparing perturbations within the same category, we find that while their RR values are comparable, their WR and LR can differ significantly.
If the WR exceeds the LR, more instances are corrected than misanswered after introducing perturbations. This suggests that
\textbf{not every spurious feature is harmful and they can even be beneficial sometimes}. 



\subsection{Further Analysis}
\label{sec:further_analysis}

The raw synthetic dataset is not ideal for extensive evaluation due to its large size. Furthermore, the class imbalance result in unfair comparisons across different types of spurious features. To facilitate more efficient evaluation, we extract the most challenging data from our synthetic datasets to create a lightweight benchmark: \textit{SIG\_Wiki} (\textbf{S}purious features \textbf{I}n \textbf{G}olden document from \textbf{Wiki}) \footnote{Specifically, we randomly select 100 instance pairs for each perturbation where both models lack robustness.}.

\begin{figure}[h]
    \centering
    \resizebox{0.45\textwidth}{!}{\includegraphics{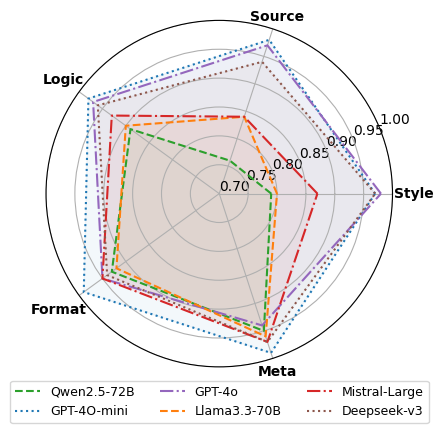}}
    \caption{Robustness comparison of six SOTA LLMs.}
    \label{fig:radar}
\end{figure}

 
\paragraph{Spurious Features in Different Model Families.}
To examine whether spurious features are merely artifacts of specific model choices, we evaluate a diverse set of SOTA LLMs on the \textit{SIG\_Wiki} benchmark.
The evaluated models include \textit{GPT-4O}, \textit{GPT-4O-mini}, \textit{Mistral-Large-Instruct} \footnote{https://huggingface.co/mistralai/Mistral-Large-Instruct-2411}, \textit{Llama-3.3-70B-Instruct}, \textit{Qwen2.5-72B-Instruct}, and \textit{DeepSeek-V3} (671B,MoE), covering a wide range of model series and architectures.
To better compare the robustness of different models, we average the RR of each perturbation within a category to derive the overall robustness for a specific type of spurious feature. The performance of six SOTA LLMs is then visualized using a radar chart, as shown in Figure \ref{fig:radar}. Despite the impressive robustness of closed-source models, they still exhibit sensitivity to certain specific perturbations.
For instance, GPT-4o achieved only an 89\% robustness rate on the datasource(twitter) perturbation.
These results demonstrate that spurious features are a widespread issue across different model families, sizes, and architectures (Dense VS. MoE).

\paragraph{Spurious Features Across Retrieval Settings.}
To evaluate whether spurious features exist across different retrieval configurations (query sources, retrievers, and evidence sources), we additionally construct a benchmark, \textit{SIG\_Trivial}, following the \textit{SURE} framework pipeline. This benchmark employs a production-level retriever (Bing Search) to retrieve documents from the open web, using queries from Trivial\_QA~\cite{joshi2017triviaqa}.
Further details are provided in Appendix~\ref{appendix:sig_trivial}. Table~\ref{tab:sure_trivial_results} presents the average RR for each type of spurious feature. The results demonstrate that spurious features are a prevalent issue for RAG systems across different retrieval configurations.

\begin{table}[ht]
\centering
\resizebox{0.48\textwidth}{!}{
\begin{tabular}{lrrrrr}
\toprule[2pt]
\textbf{} & \textbf{Style} & \textbf{Source} & \textbf{Logic} & \textbf{Format} & \textbf{Meta} \\ \addlinespace[2pt]
\hline \addlinespace[2pt]
\textbf{Mistral-7B-Instruct} & 88.0 & 94.0 & 94.5 & 94.0 & 99.0 \\
\textit{with LLM-as-Judge} & 90.5 & 91.5 & 92.0 & 93.8 & 96.0 \\
\addlinespace[2pt]  \hdashline[1pt/1pt]  \addlinespace[2pt]  
\textbf{Llama-3.1-8B-Inst.} & 87.5 & 93.5 & 93.0 & 90.8 & 97.0 \\
\textit{with LLM-as-Judge} & 85.0 & 92.0 & 91.0 & 90.8 & 93.3 \\
\bottomrule[2pt]
\end{tabular}}
\caption{Robustness evaluation of two models on \textit{SIG\_Trivial}, including results from \textit{LLM-as-Judge} paradigm as a complementary assessment.}
\label{tab:sure_trivial_results}
\end{table}

\paragraph{Reliability of Robustness Evaluations.}
To validate the reliability of our string-based evaluation method, we introduce the \textit{LLM-as-Judge} paradigm, using \textit{Llama-3.1-70B-Instruct} as the judge model, and perform a comparative assessment on \textit{SIG\_Trivial} benchmark.
See more details in Appendix~\ref{appendix:judge}.
As illustrated in Table~\ref{tab:sure_trivial_results}, our evaluation results similar to those of \textit{LLM-as-Judge}, suggesting its effectiveness and efficiency.

\paragraph{How Spurious Features Shift Model Attention.}
Interpretability provides a principled framework for linking internal states to model behaviors~\cite{zhang2026locate}. To understand why spurious features affect model predictions, we investigate how they influence the model’s internal retrieval of answer-relevant information from an attention-based perspective~\cite{liu2025mudaf}.

Specifically, we use the final token of the question $t_q$ as the query vector. For each token $t_i$ in the ground-truth answer span of the document, we extract the attention weight $\alpha(t_q, t_i)$. We then compute the average attention assigned to the ground-truth span for the original document and the perturbed document, denoted as $A_{\text{org}}$ and $A_{\text{pert}}$, respectively. The attention difference is defined as:
\begin{equation}
\Delta A = \left| A_{\text{org}} - A_{\text{pert}} \right|.
\end{equation}

We classify each example into three groups based on prediction changes after introducing spurious features: \textbf{Robust} (unchanged), \textbf{Win} (incorrect$\rightarrow$correct), and \textbf{Lose} (correct$\rightarrow$incorrect). This grouping enables us to assess whether attention shifts correlate with model's prediction changes.
To ensure a fair comparison, we randomly sample 50 examples with \textit{Golden} documents from each category in the \textit{SIG\_Wiki} benchmark, and compute the average $\Delta A$ for each group. Results are shown in Table~\ref{tab:attention_analysis}.

\begin{table}[ht]
\centering
\resizebox{0.48\textwidth}{!}{
\begin{tabular}{lrrr}
\toprule[2pt]
\textbf{Group} & $A_{\text{org}}$ & $A_{\text{pert}}$ & $\Delta A$ \\ \addlinespace[2pt]
\hline \addlinespace[2pt]
\textbf{Robust} & 4.03e-4 & 3.82e-4 & 6.52e-5 \\
\textbf{Win} & 4.65e-4 & 4.79e-4 & 9.94e-5 \\
\textbf{Lose} & 5.13e-4 & 4.84e-4 & 1.15e-4 \\
\bottomrule[2pt]
\end{tabular}}
\caption{Attention analysis on \textit{SIG\_Wiki}. $A_{\text{org}}$, $A_{\text{pert}}$, and $\Delta A$ are averaged over all examples.}
\label{tab:attention_analysis}
\end{table}

We observe that both \textbf{Win} and \textbf{Lose} cases exhibit larger attention shifts compared to \textbf{Robust} cases. Notably, in \textbf{Lose} cases, the attention assigned to the ground-truth answer within the perturbed document shows a clear decrease compared to the original.

A Welch’s t-test comparing $\Delta A$ between the \textbf{Robust} group and the Change group (Win+Lose) yields a statistically significant difference ($p = 0.046$), suggesting that internal attention variation is associated with output changes. This attention-based analysis helps explain how spurious features lead to unrobust behavior in RAG systems.

\subsection{Mitigating Spurious Features}
\label{sec:mitigate}
\begin{figure}[h]
    \centering
    \resizebox{0.48\textwidth}{!}{\includegraphics{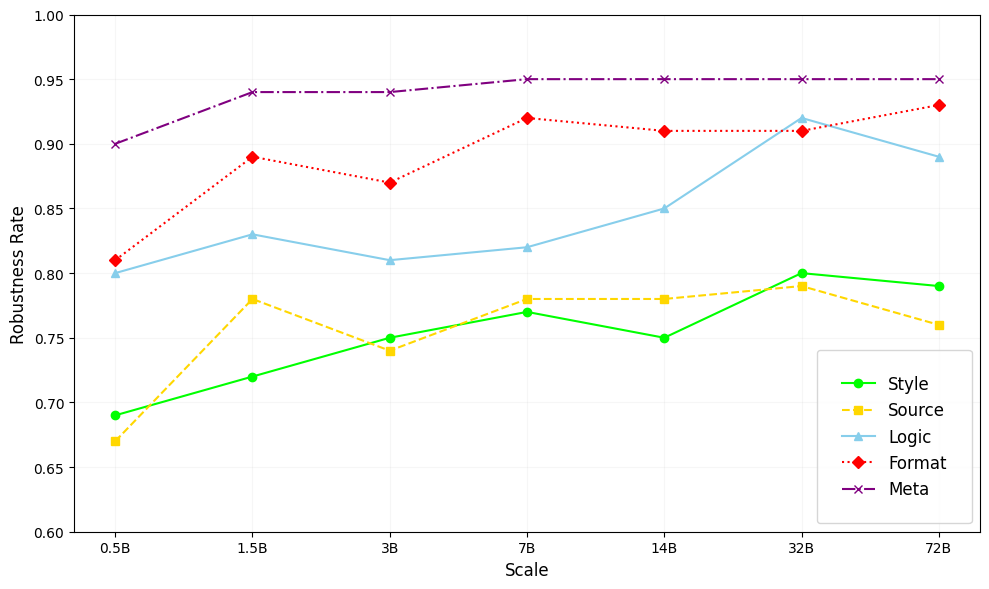}}
    \caption{Scaling analysis on Qwen2.5 series.}
    \label{fig:scale}
\end{figure}

\paragraph{Can Scaling up Model Size Solve the Problem?}
To investigate the impact of parameter scale on RAG robustness, we gradually increase the size of LLM-based readers (Qwen2.5 series, ranging from 0.5B to 72B) and evaluate their robustness across five types of spurious features. As illustrated in Figure \ref{fig:scale}, the robustness rate for all spurious features shows a relatively upward trend as the model size increases. However, when we further scale the model from 32B to 72B, the RR undergoes a significant decline (except for format and meta). 
Interestingly, for meta perturbations, while RALMs demonstrate strong robustness across all scales, their performance receives little to no benefit from scaling up. 
These findings suggest that although scaling up model size can enhance robustness to some extent, it fails to fundamentally eliminate sensitivity to spurious features.



\paragraph{Are Existing Prompting Methods Effective?}
We evaluate whether methods developed to improve the robustness of RALMs against explicit noise can generalize to spurious features. Previous work, such as Chain-of-Note (CON)~\cite{yu2023chain}, aims to enhance robustness by generating thorough rationale before producing the answer. Moreover, recent breakthroughs in the reasoning capabilities of LLMs have significantly advanced the cutting edge of RAG. By integrating with reasoning models, RAG can overcome previous limitations and adapt to more complex scenarios~\cite{gao2025synergizing}. Therefore, we test both CON and DeepSeel-R1 on our SIG benchmark (results shown in Table~\ref{tab:CON_R1}).
Notably, the robustness rate of CON is even lower than the baseline (\textit{Qwen2.5-72B-Instruct}) without applying CON. A similar phenomenon was observed in experiments with the reasoning model DeepSeek-R1~\cite{guo2025deepseek}, whose robustness was even worse than its base model, DeepSeek-V3. This indicates that the robustness against spurious features cannot be effectively improved through COT-style techniques.
\begin{table}[ht]
\centering
\resizebox{0.48\textwidth}{!}{
\begin{tabular}{lrrrrr}
\toprule[2pt]
\textbf{} & \textbf{Style} & \textbf{Source} & \textbf{Logic} & \textbf{Format} & \textbf{Meta} \\ \addlinespace[2pt]
\hline \addlinespace[2pt]
\textbf{Qwen2.5-72B} & 78.5 & 76.0 & 88.6 & 92.5 & 95.0 \\
\textbf{+ Chain-of-Note} & 74.0 & \textbf{81.7} & 66.7 & 84.8 & 91.0 \\
\addlinespace[2pt]  \hdashline[1pt/1pt]  \addlinespace[2pt]  
\textbf{DeepSeek-V3} & 96.5 & 93.6 & 95.6 & 94.0 & 96.5 \\
\textbf{DeepSeek-R1} & 84.5 & 87.3 & 83.3 & 87.0 & 87.5 \\
\bottomrule[2pt]
\end{tabular}}
\caption{Robustness evaluation of CoN and DeepSeek-R1. Values that show improvements over the baseline are marked in \textbf{bold}.}
\label{tab:CON_R1}
\end{table}

\paragraph{Improving Robustness Using Synthetic Data.}
Using the data generated by our \textit{SURE} framework, we introduce two training-based mitigation strategies, Supervised Finetuning (SFT) and Direct Preference Optimization (DPO)~\cite{rafailov2023direct}, designed to enhance the robustness of RALMs against spurious features.
Specifically, we select unrobust instances, each comprising a query, a correct answer, an incorrect answer, and two golden passages (original and perturbed documents). 
For SFT, we train the model to consistently produce the correct answer for each query, creating two training samples by pairing the answer separately with the original and perturbed golden passages.
For DPO, we train the model to prefer the correct answer over the incorrect one, likewise constructing two samples per query using both golden passages.
The goal of these strategies is to teach the model to reliably generate the correct answer, regardless of whether the input documents contain spurious features.
We train our methods for two epochs on a dataset of over 30k samples, and evaluate their robustness rates on the \textit{SIG\_Wiki} and \textit{SIG\_Trivial} (out-of-domain) benchmarks. The backbone model is \textit{Llama-3.1-8B-Instruct}. The details of the hyperparameter settings are presented in Appendix~\ref{appendix:train_setup}.

As shown in Table~\ref{tab:sft_dpo}, both methods significantly enhance robustness. However, SFT performs better on in-domain data (i.e., \textit{SIG\_Wiki}), while DPO generalizes more effectively to out-of-domain data. These results demonstrate the effectiveness of our \textit{SURE} framework for generating training data and provide strong baselines for future research.

\begin{table}[ht]
\centering
\resizebox{0.48\textwidth}{!}{
\begin{tabular}{lrrrrr}
\toprule[2pt]
\textbf{} & \textbf{Style} & \textbf{Source} & \textbf{Logic} & \textbf{Format} & \textbf{Meta} \\ \addlinespace[2pt]
\hline \addlinespace[2pt]
\textbf{Llama3.1-8B (Wiki)} & 10.0 & 15.5 & 20.0 & 24.0 & 94.0 \\
\textbf{+ SFT} & 96.5 & 94.5 & \textbf{99.0} & \textbf{99.5} & \textbf{99.7} \\
\textbf{+ DPO} & 96.5 & \textbf{96.0} & 96.0 & 98.0 & 98.0 \\
\addlinespace[2pt]  \hdashline[1pt/1pt]  \addlinespace[2pt]  
\textbf{Llama3.1-8B (Trivial)} & 87.5 & 93.5 & 93.0 & 90.8 & 97.0 \\
\textbf{+ SFT} & 88.5 & 91.5 & 95.0 & \textbf{96.3} & \textbf{99.0} \\
\textbf{+ DPO} & \textbf{94.5} & \textbf{94.5} & \textbf{97.3} & 95.8 & 98.0 \\
\bottomrule[2pt]
\end{tabular}}
\caption{Robustness comparison of \textit{Llama-3.1-8B-Instruct} trained with SFT and DPO. The upper section shows the results on \textit{SIG\_Wiki}, while the lower section presents the results on \textit{SIG\_Trivial}. The best score for each type is highlighted in \textbf{bold}.}
\label{tab:sft_dpo}
\end{table}

\section{Conclusion}
\label{sec:conclusion}
In this work, we formally highlight the spurious features problem in RAG system.
To quantify the robustness of RALMs against spurious features, we propose a novel evaluation framework, \textit{SURE}, which includes a comprehensive taxonomy, a data synthesis pipeline, and evaluation metrics. Furthermore, leveraging the synthetic data generated by \textit{SURE}, we introduce two training-based approaches that effectively improve the robustness of RALMs.
Overall, our framework enables the systematic evaluation and mitigation of spurious features, paving the way for future research.

\section*{Limitations}
\label{sec:limitation}
We strive to comprehensively cover all types of spurious features that may arise in RAG scenarios. However, some unidentifiable spurious features may fall outside the scope of our taxonomy and thus fail to be quantified using the proposed \textit{SURE} framework. 

\section*{Ethical Considerations}
\label{sec:ethics}
All datasets used in our work are publicly released under open licenses, and all models employed for generation are open-source and licensed for research use. We strictly follow the licensing terms and usage policies of these resources.

\nocite{zhang2025shifcon}
\nocite{jiang2025drp}
\bibliography{custom}

\clearpage
\section*{Appendix}
\label{sec:appendix}
\appendix

\definecolor{red-title}{RGB}{220, 20, 60} 
\definecolor{blue-back}{RGB}{230, 230, 250}    
\definecolor{blue-title}{RGB}{70, 130, 180}    
\definecolor{green-back}{RGB}{240, 255, 240}     
\definecolor{green-title}{RGB}{50, 205, 50}      
\definecolor{orange-title}{RGB}{255, 165, 0}      
\definecolor{orange-back}{RGB}{255, 228, 181}     
\definecolor{purple-back}{RGB}{238, 211, 238}     
\definecolor{purple-title}{RGB}{218, 112, 214}    

\section{Preliminary}
\label{appendix: preliminary}
In this section, we first define causal and spurious features in the context of RAG and then demonstrate the existence of spurious features statistically.



\subsection{Causal and Spurious Features in RAG}
In general, causal features are input features that have a direct causal effect on the output of predictive model~\cite{causalfeature}. Their relationship is rooted in causality, rather than mere statistical correlation. When it comes to Large Language Models, the meaning and intent of prompts serve as causal features that directly influence the models' responses. In the context of RAG, causal features refer to the semantic information of grounding data.

In contrast, spurious features are input features that co-occur with causal features and are erroneously captured by the model~\cite{spuriousfeature}. These features exhibit a statistical correlation with the model's output but lack a causal relationship. Recent research has shown that LLMs are sensitive to seemingly trivial features like prompt formatting, thereby extending the definition of spurious features to LLMs~\cite{quantifyingsensitivity}. Similarly, we define the semantic-agnostic features of the grounding data as spurious features in RAG systems.
However, conclusions drawn from in-context learning scenarios (e.g., classification and multiple-choice tasks) may not applicable to RAG scenarios, which typically involve open-ended generation tasks. Therefore, we design a preliminary experiment to validate the presence of spurious features in RAG.



\subsection{Preliminary Experiment}
\label{subsec:pre_exp}
We aim to demonstrate the semantic-agnostic features within real documents are spurious features, i.e., to reveal their impact on the output of RAG systems. 

There are some challenges in revealing the influence of semantic-agnostic features. First, when retrieving from a single corpus, it is difficult to mine semantically equivalent counterparts that differ only in semantic-agnostic features. To mine appropriate documents, we introduce \textit{Contriever-msmarco}, a traditional dense retriever, to recall 100 semantically similar candidates. To further eliminate the effect of causal features, documents without golden answers are filtered out, ensuring that the remaining documents have roughly consistent causal features. 

Still, the differences in spurious features among these candidate documents are often minor, and their impact on model responses cannot be effectively captured using binary evaluation methods that simply judge whether an answer is correct or incorrect.  
Thus, more fine-grained metrics are required to detect such nuanced performance changes. Inspired by the use of LLMs as supervision signals for document utility \cite{atlas, similarity}, we introduce the \textit{oracle score}, which measures fine-grained performance through calculating the log probability of generating correct answers given a specific document.
The \textit{oracle score} is defined as follows:
\begin{equation}
\text{Oracle}(x, y, \lower0.2ex\hbox{$\theta$}) = \sum_{t=1}^{T} \log p(y_t \mid x, y_{<t}, \lower0.2ex\hbox{$\theta$})
\end{equation}
where \(x\) is the input prompt for RALMs, including the instruction \(I\), grounding data \(G\), and query \(Q\); \(y\) represents the ground truth answer; \(\theta\) denotes the model parameters; and \(T\) is the total length of the answer sequence \footnote{For cases with multiple answers, we compute the final score by averaging the corresponding oracle scores across all answers.}. 

For each query, we construct document pairs by selecting the first-ranked and last-ranked candidate documents based on their oracle scores.
However, the presence of various semantic-agnostic features within each document pairs makes it challenging to isolate the impact of any individual features. To assess the influence of a given feature, we compare its distribution between document sets with first- and last-ranked oracle scores. A control group is constructed by randomly sampling two document sets. If the distributions differ significantly, it suggests that RALMs are sensitive to the feature. See Appendix \ref{appendix: preliminary} for implementation details of preliminary experiments. 

We test the following features: 1) Flesh Score, 2) Distinct-1, 3) Dependency Tree Depth, 4) PPL, and 5) Token Length. 
The results show that RALMs are sensitive to semantic-agnostic features. 
Nevertheless, it does not offer empirical evidence or quantitative analysis. Inspired by previous data synthesis studies~\cite{tan2024large}, we use a data synthesis approach to better control feature variables and quantify the robustness of RALMs.

\subsection{Preliminary Experiment Results}
Following the procedure described in \S\ref{subsec:pre_exp}, we use queries from the NQ-open dataset and recall candidate documents from the Wikipedia dump. After filtering, the first-ranked and last-ranked documents by oracle score yield two sets of 2,658 samples each. We employ the Kolmogorov-Smirnov (K-S) test to evaluate whether the feature distributions of the two sets are significantly different. The K-S test is a non-parametric test whose null hypothesis is that both samples are drawn from the same distribution. We reject the null hypothesis when $p < 0.05$. The following semantic-agnostic features are measured:
\begin{itemize}
\item \textbf{Flesch Score}: A readability metric designed to evaluate text difficulty. It is calculated based on the average number of syllables per word and the average number of words per sentence. The Flesch score is a number on a scale from 0 to 100, where a higher score indicates that the text is easier to read.

\item \textbf{Distinct-1}: A metric used to assess the diversity of generated text. It calculates the proportion of unique words (distinct words) to the total number of words in the output. A higher Distinct-1 score indicates that the text contains a greater variety of unique words, implying more diversity in the generated content.

\item \textbf{Dependency Tree Depth (DTD)}: A syntactic complexity metric calculated by analyzing its dependency tree. Dependency Tree Depth refers to the maximum depth of a sentence's dependency parse tree. A deeper tree suggests more complex sentence structures, while a shallower tree indicates simpler syntactic constructions.

\item \textbf{Perplexity (PPL)}: A metric used for evaluating language models, measuring how well a probabilistic model predicts a given text. It reflects the uncertainty of a language model when generating sequences of words. Lower PPL values indicate better predictive performance, meaning the model assigns higher probabilities to the actual labels in the sequence.

\item \textbf{Token Length}: We compute the total number of tokens in a text as an alternative measure of text length, given that the documents in our corpus have been pre-segmented into fixed 100-word chunks. The value is model-specific and depends on the model's vocabulary.
\end{itemize}

\begin{figure*}[htbp]
    \centering
    \subfigure[Feature distributions of the experimental group]{\includegraphics[width=0.45\textwidth]{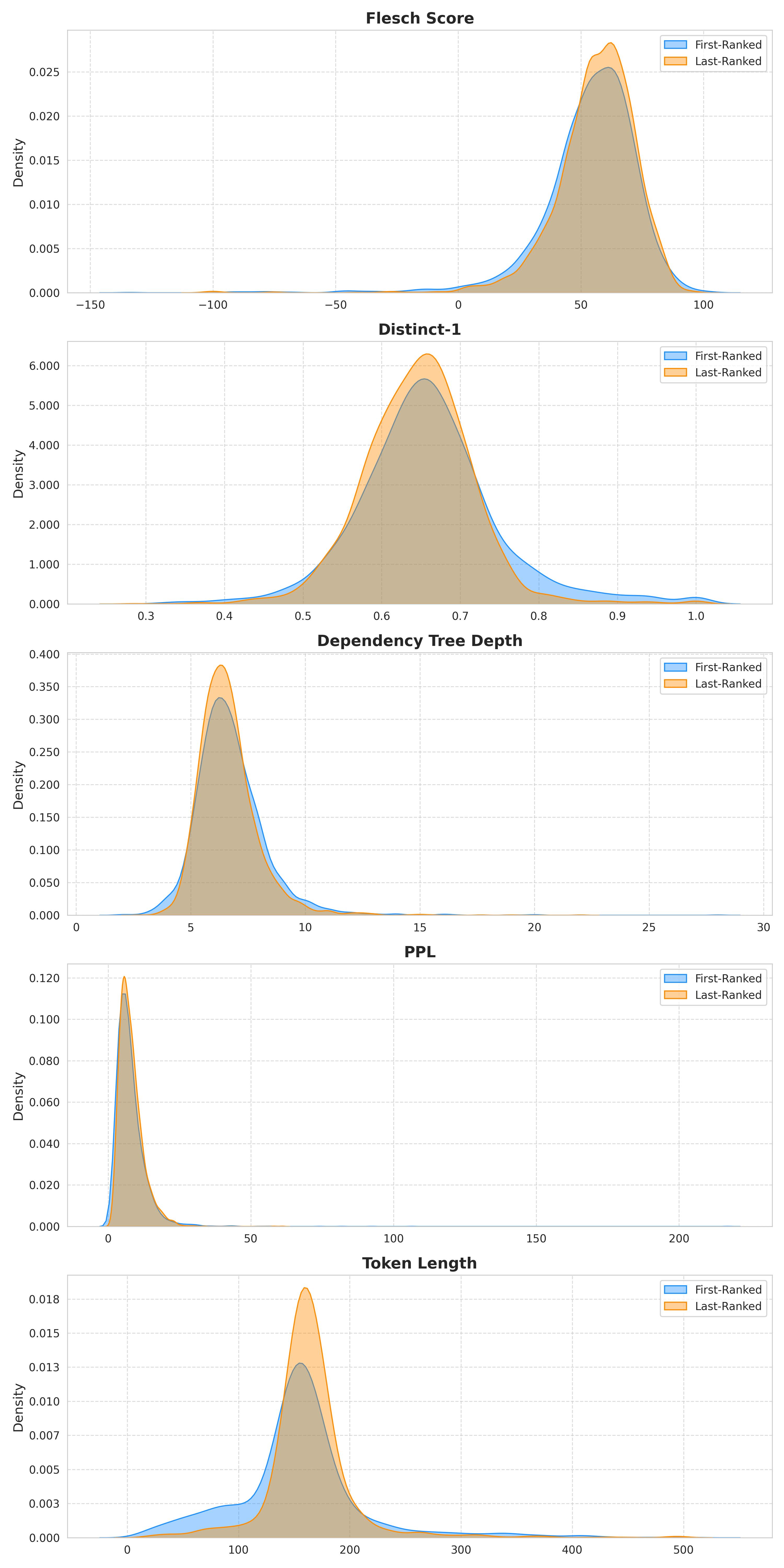}}
    \subfigure[Feature distribution of the control group]{\includegraphics[width=0.45\textwidth]{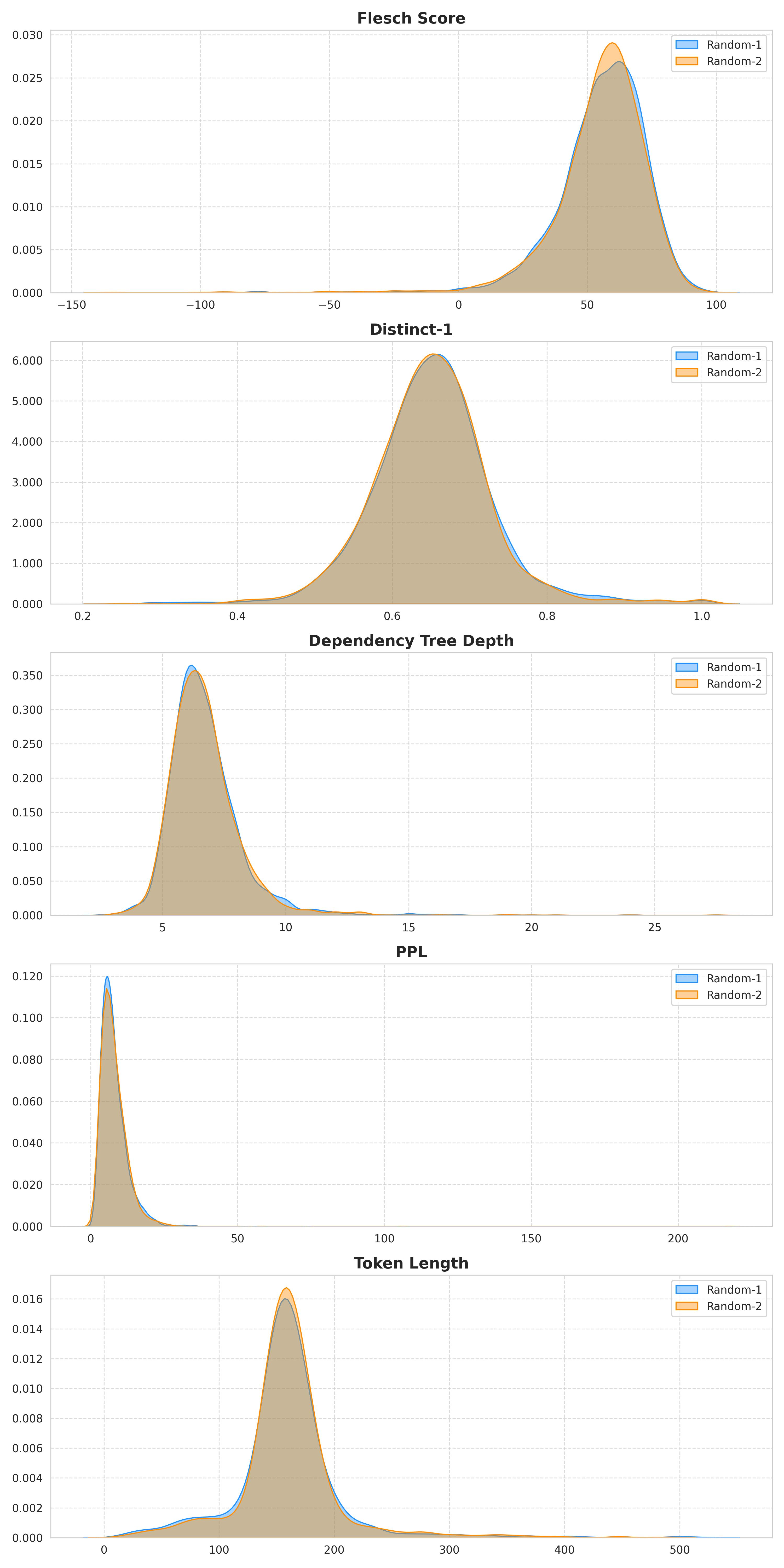}}
    \caption{Visualization of feature distributions for \textit{Mistral-7B-Instruct-v0.3}}
    \label{fig:mistral_features_distribution}
\end{figure*}

\begin{figure*}[htbp]
    \centering
    \subfigure[Feature distributions of the experimental group]{\includegraphics[width=0.45\textwidth]{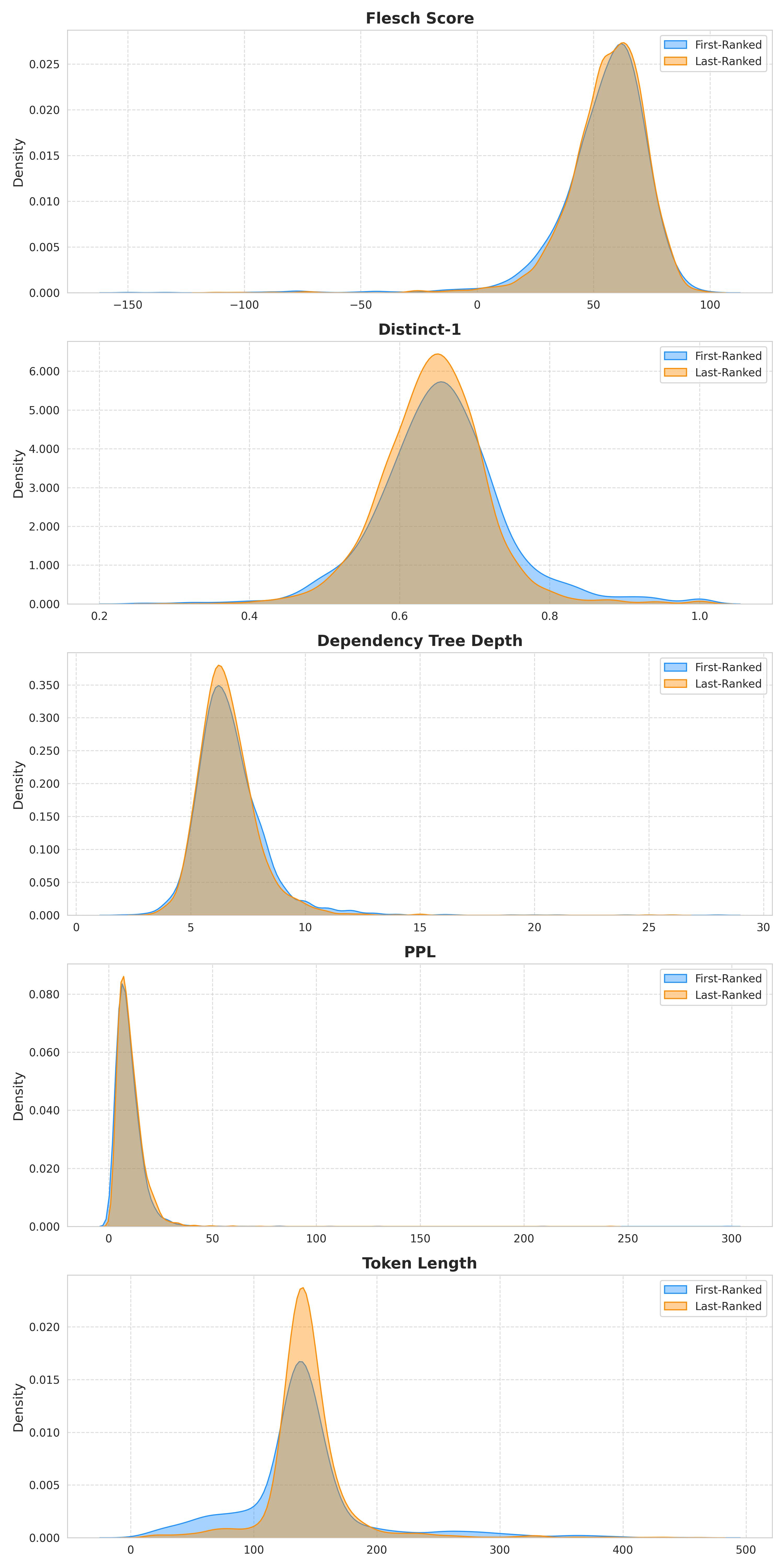}}
    \subfigure[Feature distribution of the control group]{\includegraphics[width=0.45\textwidth]{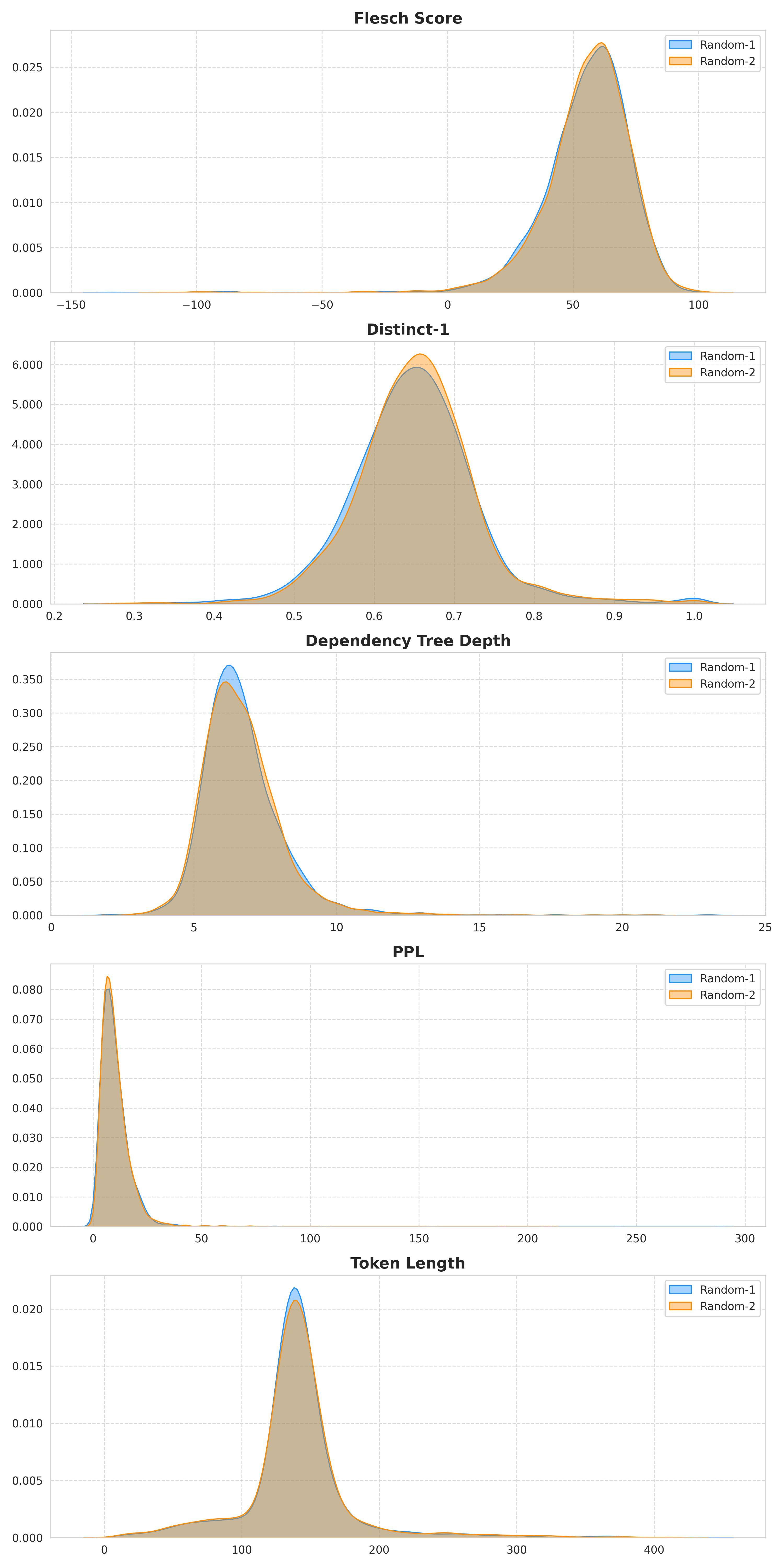}}
    \caption{Visualization of feature distributions for \textit{Llama-3.1-8B-Instruct}}
    \label{fig:llama_features_distribution}
\end{figure*}

The K-S statistic and p-values are presented in Table~\ref{tab:mistral_ks_test} and Table~\ref{tab:llama_ks_test}, with feature distributions visualized in Figure~\ref{fig:mistral_features_distribution} and Figure~\ref{fig:llama_features_distribution}. For all tested features in the experimental group, the K-S test rejects the null hypothesis, indicating significantly different distributions. In contrast, the control group fails to reject the null hypothesis. These results confirm that RALMs exhibit bias toward spurious features in documents.

\begin{table*}[htbp]
    \centering
    \begin{tabular}{lcccc}
        \toprule
        \multirow{2}{*}{} & \multicolumn{2}{c}{Experimental Group} & \multicolumn{2}{c}{Control Group} \\
        \cmidrule(r){2 - 3} \cmidrule(r){4 - 5}
        & K-S statistic & P-value & K-S statistic & P-value \\
        \midrule
        Flesch score & 0.0677 & \num{1.01e-5}$^{***}$  & 0.0301 & 0.1799\\
        Distinct-1  & 0.0756 & \num{4.95e-7}$^{***}$  & 0.0203 & 0.6431\\
        DTD & 0.0636 & \num{4.29e-5}$^{***}$ & 0.0124 & 0.9866 \\
        PPL & 0.0722 & \num{1.88e-6}$^{***}$  & 0.0162 & 0.8776 \\
        Token Length & 0.1708 &  \num{2.91e-34}$^{***}$  & 0.0256 & 0.3493 \\
        \bottomrule
    \end{tabular}
    \caption{K-S test results for \textit{Mistral-7B-Instruct-v0.3} as the oracle retriever. Significance levels: $^{**}$ $p<0.01$, $^{***}$ $p<0.001$.}
    \label{tab:mistral_ks_test}
\end{table*}

\begin{table*}[htbp]
    \centering
    \begin{tabular}{lcccc}
        \toprule
        \multirow{2}{*}{} & \multicolumn{2}{c}{Experimental Group} & \multicolumn{2}{c}{Control Group} \\
        \cmidrule(r){2 - 3} \cmidrule(r){4 - 5}
        & K-S statistic & P-value & K-S statistic & P-value \\
        \midrule
        Flesch score & 0.0305 & 0.1694 & 0.0173 & 0.8210\\
        Distinct-1  & 0.0798 & \num{8.94e-8}$^{***}$  & 0.0327 & 0.1159\\
        DTD & 0.0474 & 0.0051$^{**}$  & 0.0203 & 0.6431 \\
        PPL & 0.0538 & 0.0009$^{***}$  & 0.0181 & 0.7791 \\
        Token Length & 0.1275 &  \num{2.99e-19}$^{***}$  & 0.0188 & 0.7349 \\
        \bottomrule
    \end{tabular}
    \caption{K-S test results for \textit{Llama-3.1-8B-Instruct} as the oracle retriever. Significance levels: $^{**}$ $p<0.01$, $^{***}$ $p<0.001$.}
    \label{tab:llama_ks_test}
\end{table*}

\section{Implementation Details for Injecting Spurious Features}
\label{appendix: spurious_features_injection}
\begin{figure*}[htbp]
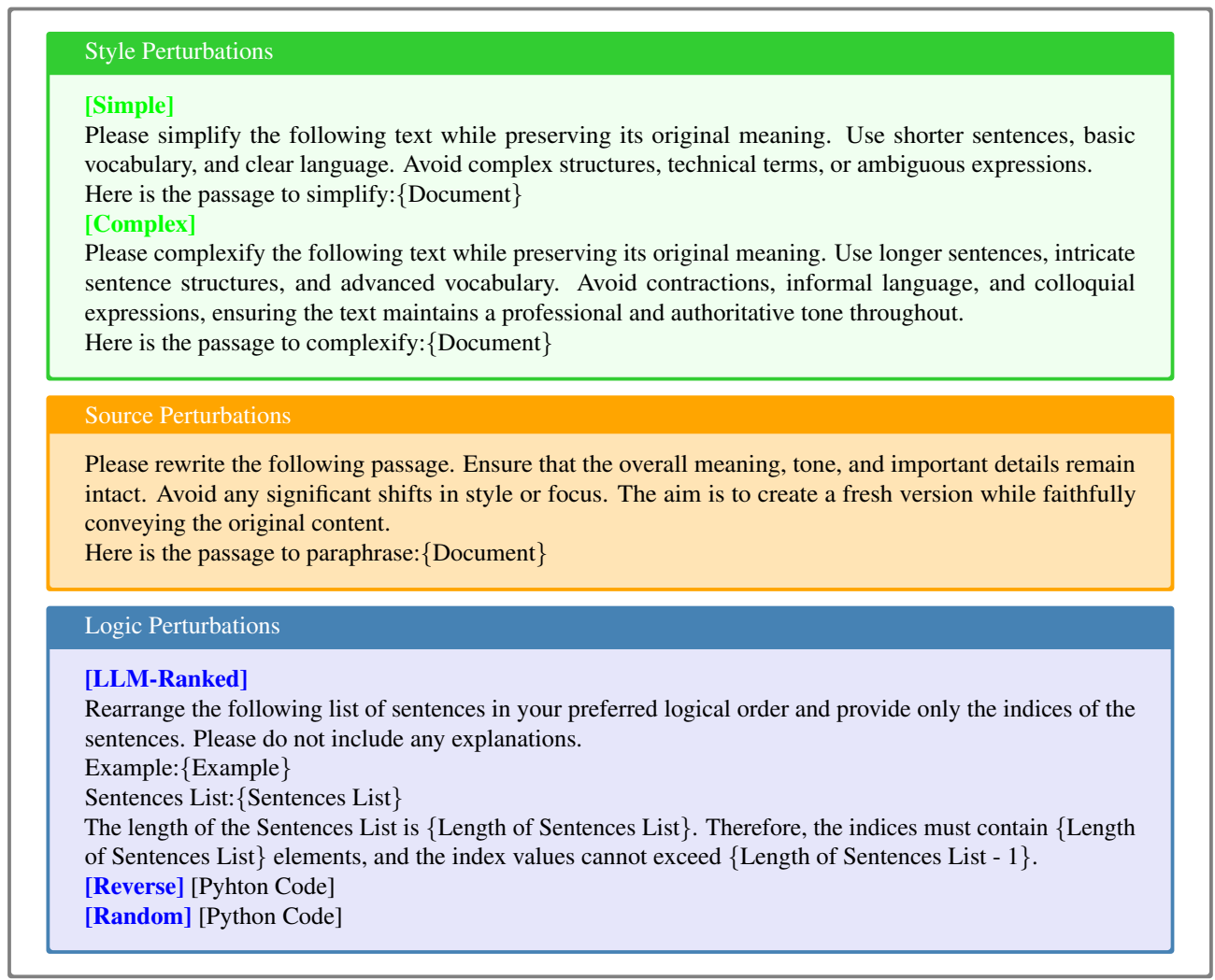

\begin{tcolorbox}[
    colback=white,
    colframe=black!50,
    boxrule=0.5mm,
    arc=0.5mm,
    outer arc=0.5mm,
]
\begin{tcolorbox}[
    boxrule=0.5mm,
    arc=0.5mm,
    outer arc=0.5mm,
    colback=green-back,
    colframe=green-title,
    title= Style Perturbations
]
\textbf{\textcolor{green}{[Simple]}}

Please simplify the following text while preserving its original meaning. Use shorter sentences, basic vocabulary, and clear language. Avoid complex structures, technical terms, or ambiguous expressions.

Here is the passage to simplify:\{Document\}

\textbf{\textcolor{green}{[Complex]}}

Please complexify the following text while preserving its original meaning. Use longer sentences, intricate sentence structures, and advanced vocabulary. Avoid contractions, informal language, and colloquial expressions, ensuring the text maintains a professional and authoritative tone throughout.

Here is the passage to complexify:\{Document\}

\end{tcolorbox}

\begin{tcolorbox}[
    boxrule=0.5mm,
    arc=0.5mm,
    outer arc=0.5mm,
    colback=orange-back,
    colframe=orange-title,
    title= Source Perturbations
]

Please rewrite the following passage. Ensure that the overall meaning, tone, and important details remain intact. Avoid any significant shifts in style or focus. The aim is to create a fresh version while faithfully conveying the original content.

Here is the passage to paraphrase:\{Document\}
\end{tcolorbox}

\begin{tcolorbox}[
    boxrule=0.5mm,
    arc=0.5mm,
    outer arc=0.5mm,
    colback=blue-back,
    colframe=blue-title,
    title= Logic Perturbations
]

\textbf{\textcolor{blue}{[LLM-Ranked]}}

Rearrange the following list of sentences in your preferred logical order and provide only the indices of the sentences. Please do not include any explanations.\\
Example:\{Example\} \\
Sentences List:\{Sentences List\} \\
The length of the Sentences List is \{Length of Sentences List\}. Therefore, the indices must contain  \{Length of Sentences List\} elements, and the index values cannot exceed  \{Length of Sentences List - 1\}.

\textbf{\textcolor{blue}{[Reverse]}}  [Pyhton Code]

\textbf{\textcolor{blue}{[Random]}}   [Python Code]
\end{tcolorbox}
\end{tcolorbox}
\caption{Prompt templates for LLM-based perturbations.}
\label{fig:llm-based_perturbations}
\end{figure*}

\begin{figure*}[htbp]
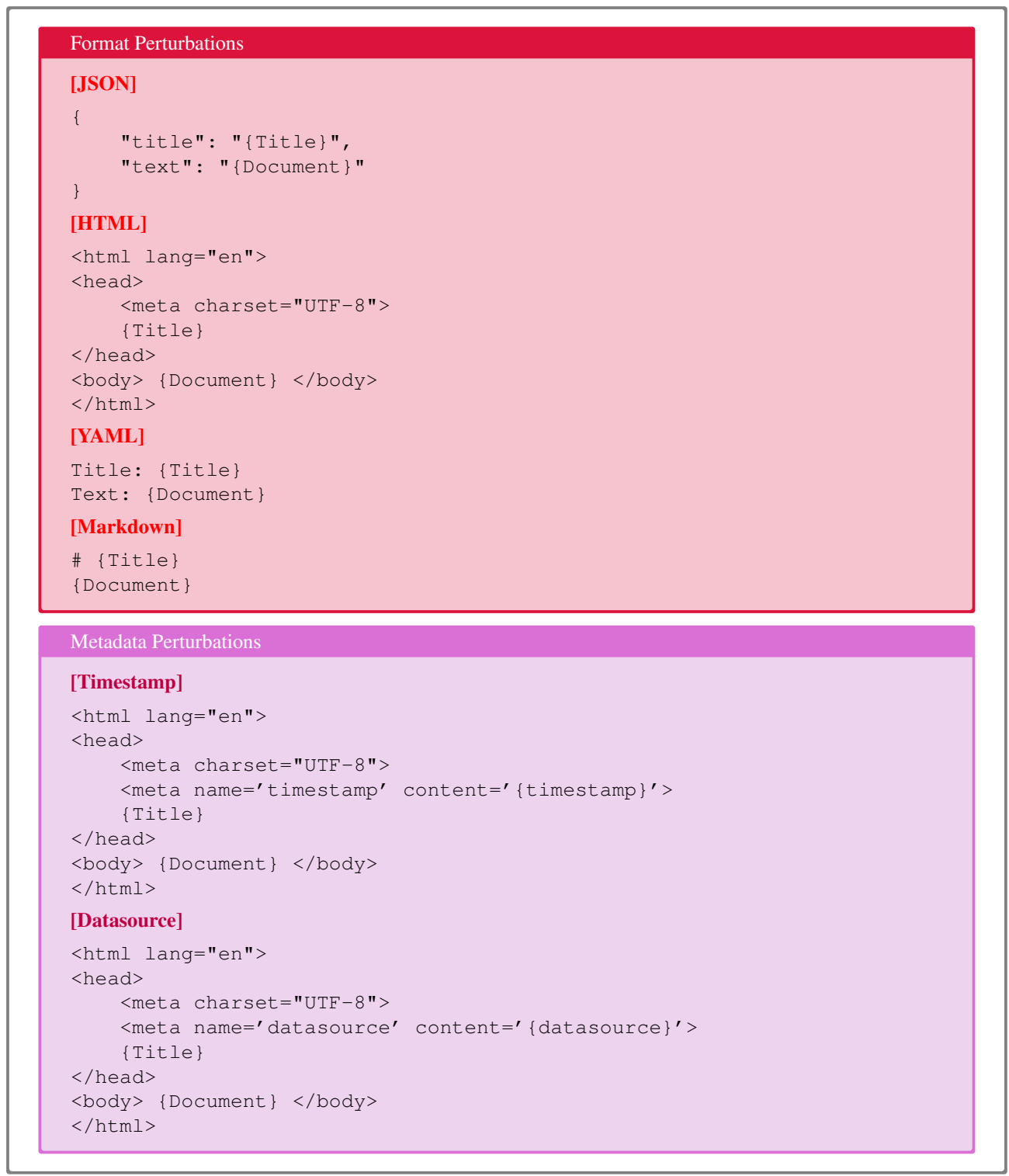

\begin{tcolorbox}[
    colback=white,
    colframe=black!50,
    boxrule=0.5mm,
    arc=0.5mm,
    outer arc=0.5mm,
]
\begin{tcolorbox}[
    boxrule=0.5mm,
    arc=0.5mm,
    outer arc=0.5mm,
    colback=red-title!25,
    colframe=red-title,
    title= Format Perturbations]
\textbf{\textcolor{red}{[JSON]}}
\begin{verbatim}
{
    "title": "{Title}",
    "text": "{Document}"
}
\end{verbatim}
\textbf{\textcolor{red}{[HTML]}}
\begin{verbatim}
<html lang="en">
<head>
    <meta charset="UTF-8">
    {Title}
</head>
<body> {Document} </body>
</html>
\end{verbatim}
\textbf{\textcolor{red}{[YAML]}}
\begin{verbatim}
Title: {Title}
Text: {Document}
\end{verbatim}
\textbf{\textcolor{red}{[Markdown]}}
\begin{verbatim}
# {Title}
{Document}
\end{verbatim}
\end{tcolorbox}

\begin{tcolorbox}[
    boxrule=0.5mm,
    arc=0.5mm,
    outer arc=0.5mm,
    colback=purple-back,
    colframe=purple-title,
    title= Metadata Perturbations
]
\textbf{\textcolor{purple}{[Timestamp]}}
\begin{verbatim}
<html lang="en">
<head>
    <meta charset="UTF-8">
    <meta name='timestamp' content='{timestamp}'>
    {Title}
</head>
<body> {Document} </body>
</html>
\end{verbatim}
\textbf{\textcolor{purple}{[Datasource]}}
\begin{verbatim}
<html lang="en">
<head>
    <meta charset="UTF-8">
    <meta name='datasource' content='{datasource}'>
    {Title}
</head>
<body> {Document} </body>
</html>
\end{verbatim}
\end{tcolorbox}
\end{tcolorbox}
\caption{Placeholder templates for rule-Based perturbations.}
\label{fig:rule-based}
\end{figure*}

We provide detailed prompts for LLM-based perturbations in Figure \ref{fig:llm-based_perturbations}. For rule-based perturbations, placeholder templates are presented in Figure \ref{fig:rule-based}.

\section{Implementation Details for Preserving Causal Features}
\label{appendix: causal_features_preservation}

We employ a bidirectional entailment algorithm to ensure the semantic equivalence before and after introducing spurious features. The prompts for its core component, NLI model, are shown in Figure \ref{fig:nli_prompts}. Furthermore, we apply a simple string-matching strategy to preserve ground truths. Specifically, for \textit{Golden} documents that originally contained the correct answers, we keep them only if they preserve the ground truths after perturbation. For \textit{Noise} documents that initially lack the correct answers, we discard them if they unexpectedly acquire ground truths due to perturbations.

\begin{figure*}[htbp]
\begin{tcolorbox}[
    colback=blue-back,
    boxrule=0.5mm,
    arc=0.5mm,
    outer arc=0.5mm]
Consider the two passages below. \\Premise: \{raw text\} \\Hypothesis: \{perturbated text\}\\
Does the premise semantically entail the hypothesis? Answer with 'entailment' if they are paraphrases,'contradiction' if they have opposing meanings, or 'neutral' if they are neither. \\
Response:
\end{tcolorbox}
\caption{Prompts for LLM-based NLI system.}
\label{fig:nli_prompts}
\end{figure*}

\section{Experimental Setup Details}
\label{appendix: setup details}

\paragraph{Prompts}
The instruction \(I\) in the RAG prompt $P= (I,G,Q)$, shown in Figure \ref{fig:RAG_prompts}, is derived from~\citet{powerofnoise}, with slight modifications to better adapt to our setting. 

\paragraph{Implementation Details}
We follow the typical "retrieve-read" setting of RAG paradigm. For the retrieval module, we use \textit{Contriever-msmarco}\footnote{\url{https://huggingface.co/facebook/contriever-msmarco}}, a BERT-based dense retriever, as the default retriever. It is finetuned on the MS MARCO dataset~\cite{msmarco} after unsupervised pretraining via contrastive learning~\cite{contriever}. To optimize the efficiency of vector similarity searches, we employ the Faiss library~\cite{faiss}. For the read module, we deploy LLMs on NVIDIA A100 GPUs and accelerate inference with vllm\footnote{https://github.com/vllm-project/vllm}. We set the temperature to 0.1 to ensure stable outputs and strong reproducibility.

\section{Statistics of the Synthetic Dataset}
\label{appendix: statistics}

We present the dataset statistics for evaluating \textit{Llama-3.1-8B-Instruct} in Table \ref{tab:data_summary_llama}.

\begin{table}[ht]
\centering
\small
\begin{tabular}{lrrrrr}
\toprule[2pt]
\textbf{} & \textbf{K-G} & \textbf{K-N} & \textbf{U-G} & \textbf{U-N} & \textbf{Total} \\ \addlinespace[2pt]
\hline \addlinespace[2pt]
\textbf{Style} & 7321 & 28975 & 3038 & 39869 & 79203 \\
\textbf{Source} & 8768 & 30145 & 3709 & 41391 & 84013 \\
\textbf{Logic} & 9229 & 33294 & 4082 & 44233 & 90838 \\
\textbf{Format} & 10481 & 35616 & 4697 & 47920 & 98714 \\
\textbf{Meta} & 10563 & 35451 & 4796 & 47987 & 98797 \\
\bottomrule[2pt]
\end{tabular}
\caption{Distribution of the \textit{SURE\_Wiki} dataset for \textit{Llama-3.1-8B-Instruct}.}
\label{tab:data_summary_llama}
\end{table}

\begin{figure*}[htbp]
\begin{tcolorbox}[
    colback=blue-back,
    boxrule=0.5mm,
    arc=0.5mm,
    outer arc=0.5mm]
You are given a question and you MUST respond by EXTRACTING the answer (max 5 tokens) from the provided document. If the document does not contain the answer, respond with NO-RES.
\end{tcolorbox}
\caption{Instruction \(I\) used for the QA task.}
\label{fig:RAG_prompts}
\end{figure*}

\begin{table*}[h]
\centering
\vspace{-0.2cm}
\renewcommand{\arraystretch}{1.1}
\resizebox{1.0\textwidth}{!}{
\begin{tabular}{clcccccccccccccccc}
\toprule[2pt]
                     \multicolumn{18}{c}{\textit{Llama-3.1-8B-Instruct}}                      \\ \addlinespace[2pt]
\cline{1-18} \addlinespace[2pt]
\multirow{2}{*}{\textbf{Taxonomy}}       & \multirow{2}{*}{\textbf{Perturbations}}             & \multicolumn{5}{c}{\textbf{Known-Golden}} & \multicolumn{5}{c}{\textbf{Known-Noise}} & \multicolumn{5}{c}{\textbf{Unknown-Golden}} & \multicolumn{1}{c}{\textbf{U-N}}  \\
\cmidrule(lr){3-7} \cmidrule(lr){8-12} \cmidrule(lr){13-17} \cmidrule(lr){18-18}
&                                & LR & RR & WR & Org & Acc & LR & RR & WR & Org & Acc & LR & RR & WR & Org & Acc & RR \\ \addlinespace[2pt]
\hline \addlinespace[2pt]
\multirow{2}{*}{Style}     & Simple & 7.79        & 83.04        & \textbf{9.18}   & \multirow{2}{*}{66.03}      & 67.42     &  1.70     & 95.80      & \textbf{2.50}  &  \multirow{2}{*}{4.12}    &  4.92        &  8.43       & 82.88        &  \textbf{8.69}       & \multirow{2}{*}{51.42}   & 51.68    & 99.45       \\
                            & Complex &  6.00      &  85.60      &  \textbf{8.40}   &    
 & 68.43     &  1.91     &  96.59     & 1.50  &     & 3.71     &  6.71       &  84.86       & \textbf{8.43}  &    &  53.13        &  99.57      \\ \addlinespace[2pt]  \hdashline[1pt/1pt]  \addlinespace[2pt]
  
\multirow{2}{*}{Source}       & LLM-Generated & 5.89       & 86.43      & \textbf{7.69}   &  \multirow{2}{*}{65.62} &  67.43   & 1.43      & 96.83      & \textbf{1.74}  &  \multirow{2}{*}{4.13}   & 4.45   &  6.20      &  85.71       & \textbf{8.09}   & \multirow{2}{*}{49.15}
  & 51.04     &  99.56         \\
                            & Self-Generated & 6.55       & 85.01      & \textbf{8.44}    &    &  67.52    & 1.55   &  96.37     &\textbf{2.09}    &     &  4.67  & 6.52       &  86.36      & \textbf{7.12}     &   & 49.74      &   99.57       \\
\addlinespace[2pt]  \hdashline[1pt/1pt]  \addlinespace[2pt]  
\multirow{3}{*}{Logic}       & Reverse & 5.06       &  90.82      &  4.12 & \multirow{3}{*}{62.95}  &  62.01     & 1.13      & 97.82      & 1.06  &  \multirow{3}{*}{4.43}  &  4.36 & 5.73        & 89.71      & 4.56    & \multirow{3}{*}{45.84}   & 44.66      & 99.67       \\
                            & Random & 3.91       & 93.16       & 2.93   &  & 61.97    & 0.86      & 98.31      & 0.83   &    & 4.40   &  4.21       & 91.67        & 4.12       &      & 45.74     &  99.72       \\
                            & LLM-Ranked &  3.24      &   93.93     & 2.83  &    & 62.54     & 0.82      & 98.43      & 0.74   &    & 4.36  & 3.58        & 93.36        &3.06    &    & 45.32       &  99.76     \\
\addlinespace[2pt]  \hdashline[1pt/1pt]  \addlinespace[2pt]     
\multirow{4}{*}{Format}       & JSON &  7.01     & 88.25       & 4.74    & \multirow{4}{*}{63.91}   & 61.64   & 1.70    &  97.25    & 1.05  & \multirow{4}{*}{3.87}  & 3.21  & 5.92        &  89.63       & 4.45   &  \multirow{4}{*}{49.35}   &  47.88      & 99.61         \\
                            & HTML &  11.85      & 84.46      & 3.69   &     & 55.75   & 2.70      & 96.90     & 0.40  &    & 1.56   & 9.33        & 86.78        &  3.90     &   & 43.92       & 99.61     \\
                            & YAML &  5.26     &  89.94     & 4.80    &    & 63.45  & 1.26      & 97.41      & 1.33   &    & 3.94    & 4.79        & 90.80      & 4.41     &    & 48.97       &  99.67       \\
                            & Markdown &  2.32     & 92.23      & \textbf{5.45} &   & 67.04    & 0.60      &  96.89     &  \textbf{2.51}   &    & 5.77  & 2.34       & 93.46      & \textbf{4.19}    &    &  51.20     & 99.61          \\
\addlinespace[2pt]  \hdashline[1pt/1pt]  \addlinespace[2pt]    
\multirow{4}{*}{Metadata}      & Timestamp (pre) & 2.08       & 95.81       & \textbf{2.11}  & \multirow{4}{*}{55.77}   & 55.80    & 0.28      & 99.42      & \textbf{0.29}    &  \multirow{4}{*}{1.58}  &   1.59   & 2.54        &   95.56      &  1.90    & \multirow{4}{*}{43.31}     & 42.66       & 99.95         \\
                            & Timestamp (post) & 2.04       &  95.86      & \textbf{2.10}  &   & 55.84    & 0.25      &  99.43     & \textbf{0.32}  &    &1.64     & 2.81       & 95.56      & 1.63     &   &  42.12       & 99.95            \\    
                             & Datasource (wiki) &  2.11     & 93.45       &\textbf{4.44}   &    & 58.10      & 0.23      &  98.96     & \textbf{0.81}    &     &  2.17   & 3.25       & 92.47      & \textbf{4.27}        &       &  44.33       & 99.86          \\
                            & Datasource (twitter) & 2.27        & 94.11      & \textbf{3.62}   &     &  57.11    & 0.31     &  99.25     & \textbf{0.43}   &    & 1.70   & 2.77        &  93.97      & \textbf{3.25}    &       &  43.79      &  99.91        \\
\bottomrule[2pt]
\end{tabular}}
\caption{Robustness evaluation results of \textit{Llama-3.1-8B-Instruct} on the synthetic dataset. We use \textbf{Bold} to mark the WR values that are higher than the LR, suggesting that the perturbation is beneficial.}
\label{tab:llama3_sure_results}
\end{table*}

\section{Details of \textit{SIG\_Trivial} Benchmark}
\label{appendix:sig_trivial}
We additionally construct a new benchmark, \textit{SIG\_Trivial}, following the \textit{SURE} framework pipeline. This new benchmark uses a production-level retriever (Bing Search) to retrieve documents from the open web, with queries sources, retrieval algorithms, and data sources all differing from those used in our original SIG dataset. Specifically, we sample 1000 queries from TriviaQA and then sample 3 web documents (longer length than passages from wikepdeia) with golden answers from Bing Search for each query to serve as the grounding data. For each subcategory, 100 instances are randomly sampled to construct the new benchmark.

\section{Implementation of \textit{LLM-as-Judge}}
\label{appendix:judge}
The \textit{LLM-as-Judge} paradigm has been widely adopted as an automatic evaluation method in prior work~\cite{li2025generation,gao2023human}.
We provide two demonstrations and prompt the LLM (\textit{Llama-3.1-70B-Instruct}) to act as an evaluator. Before reaching a final judgment, the model is instructed to reason step by step using the chain-of-thought (COT) approach, encouraging a more thorough consideration of answer quality. The prompt template is shown in Figure~\ref{fig:judge_prompt}.

\section{Training Setups}
\label{appendix:train_setup}
We perform full parameter fine-tuning for both SFT and DPO using the AdamW optimizer with DeepSpeed ZeRO-3 on eight NVIDIA A100-SXM4-80GB GPUs. The detailed hyper-parameters are listed in Table~\ref{tab:hyperparameters}.

\begin{table}[!h]
\centering
\renewcommand{\arraystretch}{1.2}
\begin{tabular}{@{} l c c @{}}
\toprule[1.5pt]\toprule[1pt]
\textbf{Hyper-parameter} & \textbf{SFT} & \textbf{DPO} \\
\midrule
Learning Rate & $1 \times 10^{-5}$ & $5 \times 10^{-6}$ \\
Number of Epochs & 2 & 2 \\
Per-device Batch Size & 8 & 1 \\
Gradient Accumulation Steps & 1 & 2 \\
Effective Batch Size & 64 & 16 \\
Learning Rate Scheduler & cosine & cosine \\
Warmup Ratio & 0.1 & 0.1 \\
Max Sequence Length & 8192 & 8192 \\
\bottomrule[1pt]\bottomrule[1.5pt]
\end{tabular}
\caption{Hyper-parameters for SFT and DPO training.}
\label{tab:hyperparameters}
\end{table}

\end{document}